%% file: paper.tex
\documentclass[10pt,twocolumn,letterpaper]{article}

\usepackage[pagenumbers]{cvpr} 
\usepackage[accsupp]{axessibility}

\usepackage{times}
\usepackage{epsfig}
\usepackage{graphicx}
\usepackage{amsmath}
\usepackage{color}
\usepackage{amssymb}
\usepackage{lipsum}
\usepackage{algorithm}
\usepackage{algpseudocode}
\usepackage{xcolor}
\usepackage{tabularx}
\usepackage{multirow}
\usepackage{enumitem}
\usepackage{bbm}
\usepackage{wrapfig}
\usepackage{setspace}
\usepackage{footnote}
\usepackage{afterpage}

\usepackage{booktabs}
\usepackage{colortbl} 
\usepackage[breaklinks=true,colorlinks,bookmarks=true]{hyperref}

\newcommand\blfootnote[1]{
  \begingroup
  \renewcommand\thefootnote{}\footnote{#1}
  \addtocounter{footnote}{-1}
  \endgroup
}

\newcommand{\Fig}[1]{Fig.~\ref{fig:#1}}

\newcommand{\Tbl}[1]{Table~\ref{tab:#1}}

\def\ie{\emph{i.e.}}
\def\eg{\emph{e.g.}}

\definecolor{bostonuniversityred}{rgb}{0.8, 0.0, 0.0}

\newcommand{\vspaceBottomFigure}{{\vspace{-3.5mm}}}
\newcommand{\vspaceBottomTable}{{\vspace{-3.0mm}}}

\def\etal{\emph{et al.}}

\begin{document}

\title{Human Pose Estimation in Extremely Low-Light Conditions}

\author{Sohyun Lee$^{1}$$^{\ast}$
\hspace{8mm}
Jaesung Rim$^{1}$$^{\ast}$
\hspace{8mm}
Boseung Jeong$^{2}$ 
\hspace{8mm}
Geonu Kim$^{2}$
\hspace{8mm} 
Byungju Woo$^{2}$ \vspace{0.5mm} \\ 
\hspace{2mm}
Haechan Lee$^{1}$ 
\hspace{8mm}
Sunghyun Cho$^{1,2}$$^{\dagger}$
\hspace{8mm}
Suha Kwak$^{1,2}$$^{\dagger}$ \vspace{2.0mm}\\
$^1$Graduate School of AI, POSTECH \qquad \ \       
$^2$Dept. of CSE, POSTECH\\
\and
\vspace{-10mm}\\
{\tt \small \url{http://cg.postech.ac.kr/research/ExLPose}}
}

\input{Paper/Figures/Teaser}

\maketitle

\input{Paper/_0_abstract}
\input{Paper/_1_introduction}
\input{Paper/_2_relatedwork}

\input{Paper/_3_dataset}
\input{Paper/_4_method}
\input{Paper/_5_experiments}
\input{Paper/_6_conclusion}

\input{Supplement/Supp}

\clearpage
{\small
\bibliographystyle{ieee_fullname}
\bibliography{paper}
}

\end{document}

%% file: Paper/Figures/Teaser.tex
\twocolumn[{%
\renewcommand\twocolumn[1][]{#1}
\maketitle
\begin{center}
\centering
\begin{minipage}{1.0\linewidth}
\vspace{-10mm}
\centering
\includegraphics[width=0.99\textwidth]{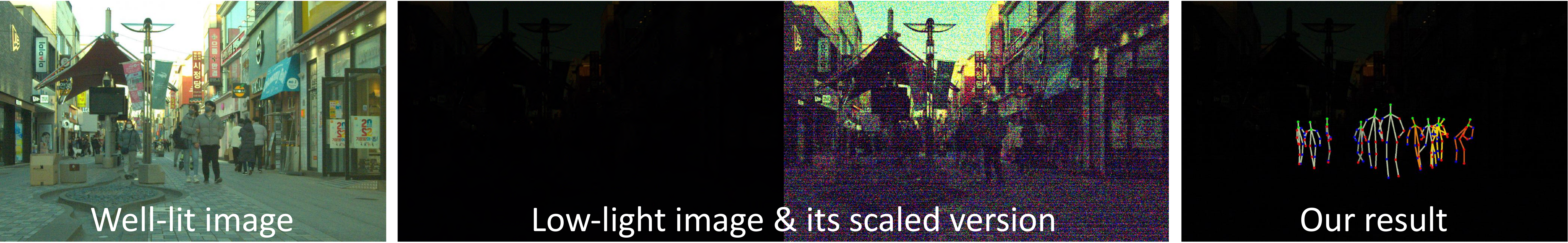}
\vspace{+1.5mm}
\label{fig:Teaser}
\end{minipage}
\end{center}
}]

%% file: Paper/_0_abstract.tex
\begin{abstract}
We study human pose estimation in extremely low-light images. This task is challenging due to the difficulty of collecting real low-light images with accurate labels, 
and severely corrupted inputs that degrade prediction quality significantly.
To address the first issue, we develop a dedicated camera system and build a new dataset of real low-light images with accurate pose labels. Thanks to our camera system, each low-light image in our dataset is coupled with an aligned well-lit image, which enables accurate pose labeling and is used as privileged information during training. We also propose a new model and a new training strategy that fully exploit the privileged information to learn representation insensitive to lighting conditions. Our method demonstrates outstanding performance on real extremely low-light images, and extensive analyses validate that both of our model and dataset contribute to the success. 
\blfootnote{$^{\ast}$ Equal contribution. $^{\dagger}$ Corresponding authors.}
\end{abstract}

%% file: Paper/_1_introduction.tex
\section{Introduction}
Deep neural networks~\cite{chen2018cascaded,sun2019deep,toshev2014deeppose,xiao2018simple,zhang2019fast} trained with large-scale datasets~\cite{MPII_pose_dataset,guler2018densepose,johnson2010clustered,li2019crowdpose,Mscoco} have driven dramatic advances in human pose estimation recently.
However, their success demands high-quality inputs taken in controlled environments while in real-world applications images are often corrupted by low-light conditions, adverse weather conditions, sensor noises, motion blur, \etc~
Indeed, a precondition for human pose estimation \emph{in the wild} is robustness against such adverse conditions.

Motivated by this, 
we study pose estimation under \emph{extremely} low-light conditions using a single sRGB image, in which humans can barely see anything.
The task is highly practical as its solution enables nighttime applications of pose estimation
without raw-RGB data or additional devices like IR cameras. 
It is at the same time challenging due to the following two reasons. 
The first is the difficulty of data collection. Manual annotation of human poses in low-light images is often troublesome due to their limited visibility.  
The second is the difficulty of pose estimation on low-light images.
The poor quality of low-light images in terms of visibility and signal-to-noise ratio largely degrades prediction accuracy of common pose estimation models.
A na\"ive way to mitigate the second issue is to apply low-light image enhancement~\cite{chen2018learning,jiang2021enlightengan,lore2017llnet,lv2018mbllen,wang2021low} to input images.
However, image enhancement is in general highly expensive in both computation and memory.
Also, it is not aware of downstream recognition tasks and thus could be sub-optimal for pose estimation in low-light conditions.

To tackle this challenging problem, we first present a new dataset of real extremely low-light images with ground-truth pose labels.
The key feature of our dataset is that each low-light image is coupled with a well-lit image of the same content.
The advantage of using the well-lit images is two-fold.
First, they enable accurate labeling for their low-light counterparts thanks to their substantially better visibility.
Second, they can be utilized as \emph{privileged information}~\cite{LUPI_SSVM+,Heteroscedastic_Dropout,LopSchBotVap_ICLR16,LUPI_2015,LUPI_2009}, \ie, additional input data that are more informative than the original ones (low-light images in our case) but available only in training, to further improve performance on low-light images.
Such benefits of paired training images have also been validated in other robust recognition tasks~\cite{dai2020curriculum,Sakaridis_2019_ICCV,Sakaridis_2018_ECCV,Sakaridis_2018_IJCV,acdc,lee2022fifo}.
The beauty of our dataset is that pairs of low-light and well-lit images are all \emph{real} and \emph{aligned}, unlike existing datasets that provide pairs of synthetic-real images~\cite{dai2020curriculum,Sakaridis_2018_IJCV,Sakaridis_2018_ECCV} or those of largely misaligned real images~\cite{Sakaridis_2019_ICCV,acdc}.
Since it is practically impossible to capture such paired images using common cameras, we build a dedicated camera system for data collection.

We also propose an effective method based on learning using privileged information (LUPI)~\cite{LUPI_2009} to fully exploit our dataset.
The proposed method considers a model taking low-light inputs as a \emph{student} and a model dealing with corresponding well-lit images as a \emph{teacher}.
Both of the teacher and student are trained by a common pose estimation loss, and the student further utilizes knowledge of the teacher as additional supervision. 
Specifically, our method employs neural styles of intermediate feature maps as the knowledge and forces neural styles of low-light images to approximate those of well-lit images by an additional loss. 
As will be demonstrated, this LUPI approach allows the learned representation to be insensitive to lighting conditions.
Moreover, we design a new network architecture that unifies the teacher and student through lighting-condition specific batch normalization (LSBN).
LSBN consists of two batch normalization (BN) layers, each of which serves images of each lighting condition, \ie, `well-lit' or `low-light'.
We replace BNs of an existing network with LSBNs so that images of different lighting conditions are processed by different BNs.
Hence, in our architecture, the teacher and student share all the parameters except for those of their corresponding BNs, which allows the student to enjoy the strong representation learned using well-lit images.

The efficacy of our method is evaluated on real low-light images we collected for testing. 
Our method outperforms its reduced versions and relevant approaches such as lighting-condition adversarial learning and a combination of image enhancement and pose estimation.
These results clearly demonstrate the advantages of our dataset and method.
In short, our major contribution is three-fold: 
\vspace{-1.5mm}
\begin{itemize}[leftmargin=5mm] 
\itemsep=-0.5mm
    \item We propose a novel approach to human pose estimation in extremely low-light conditions using a single sRGB image. To the best of our knowledge, we are the first to tackle this challenging but highly practical problem.
    \item We build a new dataset that provides real and aligned low-light and well-lit images with accurate pose labels. 
    \item We present a strong baseline method that fully exploits the low-light and well-lit image pairs of our dataset. 
\end{itemize}

%% file: Paper/_2_relatedwork.tex
\section{Related Work}

\noindent \textbf{Low-light Image Enhancement.}
Classical methods for low-light image enhancement have been developed based on histogram equalization and the Retinex theory~\cite{SSR_1997_TIP,BPDHE_2007_TCM,LDR_2013_TIP,LIME_2017_TIP}.
Recently, learning-based methods have driven remarkable advances in this field; examples include
an auto-encoder~\cite{lore2017llnet}, a multi-branch architecture~\cite{lv2018mbllen}, and a U-Net architecture~\cite{chen2018learning}. 
Also, Jiang~\etal~\cite{jiang2021enlightengan} proposed a GAN using unpaired low-light and well-lit images, and
Wang~\etal~\cite{wang2021low} learned mapping from low-light  to well-lit images via normalizing flow.
They usually require heavy computation and are learned without considering downstream recognition tasks.
In contrast, we focus on learning features insensitive to lighting-condition for pose estimation while bypassing low-light image enhancement.

\noindent \textbf{Low-light Datasets.}
SID~\cite{chen2018learning} and LOL~\cite{Chen2018Retinex} provide paired low-light and well-lit images. 
Wang~\etal~\cite{Wang_2021_ICCV} collected paired low-light and well-lit videos by playing motion using an electric slide system. 
Due to the difficulty of capturing paired images simultaneously, they only provide images of static objects.
Meanwhile, Jiang~\etal~\cite{Jiang_2019_ICCV} built a dual-camera system to capture paired low-light and well-lit videos at once.
Inspired by this, we construct a dedicated camera system for collecting paired low-light and well-lit images of humans with motion.
Low-light datasets have also been proposed for other tasks, \eg, 
ARID~\cite{xu2020arid} for human action recognition, NOD~\cite{morawski2021nod} and ExDark~\cite{loh2019getting} for object detection.
While these datasets provide only low-light images for training recognition models, our dataset provides low-light images along with their well-lit counterparts as well as accurate human pose labels.

\noindent \textbf{Pose Estimation in Low-light Conditions.}
Crescitelli~\etal~\cite{Crescitelli_SAS,Crescitelli_TIM} presented a human pose dataset containing 1,800 sRGB and 2,400 infrared (IR) images captured at night, and integrated sRGB and IR features for human pose estimation.
Our work is clearly distinct from this in two aspects. 
First, we focus on human pose estimation in \emph{extremely} low-light images, in which humans barely see anything. 
On the other hand, Crescitelli~\etal~\cite{Crescitelli_SAS,Crescitelli_TIM} captured images of night scenes with light sources, in which most human objects are sufficiently visible. 
Second, our model takes a single sRGB image captured in a low-light environment as input for pose estimation 
without demanding extra observations like IR images.
To the best of our knowledge, our work is the first to tackle this challenging but highly practical task.

\noindent \textbf{Robust Visual Recognition.}
The performance of conventional recognition models often gets degraded in adverse conditions~\cite{loh2019getting,liang2021recurrent,HLAFace_2021_CVPR,morawski2021nod}.
To address this issue, the robustness of visual recognition has been actively studied~\cite{goodfellow2014explaining,Hendrycks2019_ImageNetC,hendrycks2018deep,schneider2020improving,son2020urie,shi2020informative,acdc,lee2022fifo}.
Low-light visual recognition is one such direction that aims at learning features robust to limited visibility of low-light images.
Sasagawa~\etal~\cite{sasagawa2020yolo} utilized a U-Net architecture to restore well-lit images from raw-RGB low-light images for object detection.
Morawski~\etal~\cite{morawski2021nod} incorporated an image enhancement module into an object detector and proposed lighting variation augmentations for nighttime recognition.
However, the enhancement module is often substantially heavy to be integrated with recognition models.
We instead learn a pose estimation model insensitive to lighting conditions so that it does not need an extra module for image enhancement.

\noindent \textbf{Learning Using Privileged Information.}
LUPI aims at exploiting privileged information available only in training to improve target models in terms of accuracy, label-efficiency, and convergence speed.
Vapnik~\etal~\cite{LUPI_2009} first introduced LUPI for support vector machine classifiers.
The idea has been extended to tackle various tasks beyond classification, \eg, object localization~\cite{LUPI_SSVM+}, metric learning~\cite{Fouad_LUPI}, ranking~\cite{Sharmanska_LUPI}, and clustering~\cite{Feyereisl_LUPI_Clustering}.
LUPI has also been studied for training deep neural networks:
Lopez-Paz~\etal~\cite{LopSchBotVap_ICLR16} investigated the relation between LUPI and knowledge distillation~\cite{hinton2015distilling},
Lambert~\etal~\cite{Heteroscedastic_Dropout} developed a new dropout operation controlled by privileged information,
and Hoffman~\etal~\cite{Hoffman_sideinfo} employed an auxiliary model that approximates the teacher using ordinary input data.
Our work is clearly distinct from these in terms of the model architecture and the way of teacher-student interaction as well as the target task: Our model maximizes parameters shared by the teacher and student so that privileged information helps improve representation quality of the student, and the student is trained to approximate internal behavior of the teacher as well as predicting human poses.

%% file: Paper/_3_dataset.tex
\input{Paper/Figures/Dual_camera}

\section{ExLPose Dataset} \label{sec:dataset}
We propose the first \emph{Extremely Low-light} image dataset for human Pose estimation, coined ExLPose.
The ExLPose dataset provides pairs of low-light and well-lit images that are real and aligned, as well as their ground-truth human pose labels.
We believe that the accurate pose labels and well-lit counterparts for extremely low-light images will open a new and promising research direction towards human pose estimation in low-light conditions.

Since it is practically impossible to simultaneously capture paired low-light and well-lit images using common cameras, we construct a dual-camera system~\cite{Jiang_2019_ICCV, rim_2022_ECCV}
dedicated to the purpose.
Our camera system is depicted in \Fig{system}. 
It consists of two camera modules with a beam splitter that distributes light from a lens equally to the two camera modules. 
One of the modules captures well-lit images, and the other captures low-light images through a 1\% neutral density (ND) filter that optically reduces the amount of light by 100 times.
The two camera modules simultaneously capture a pair of low-light and well-lit images of the same scene with a synchronized shutter. 
As done in~\cite{jsrim-ECCV2020}, the images are geometrically aligned using a reference image pair captured in a static scene. Details of the geometric alignment are given in Sec.~\ref{appendix:geometric_alignment} of the supplement.

\input{Paper/Figures/ExLPose_pair}
\input{Paper/Tables/ExLPose_statistics}
\input{Paper/Figures/ExLPose_examples}

We collected images of various indoor and daytime outdoor scenes in the sRGB format. 
To cover low-light scenes of diverse brightness levels, we collected low-light images with various exposure times.
Specifically, for each scene, we first manually found a proper exposure time and a gain value to capture a well-lit image without losing much information in the highlights or in the shadows, and also without motion blur.
Then, we sequentially captured low-light images reducing the exposure time by 1, 2, 4, 8, and 12 times with the same gain value as shown at the bottom of \Fig{pair}.
At the same time, for the well-lit images, we also changed the exposure time in the same manner to synchronize the exposure time for each frame, but inverse-proportionally increased the gain value to maintain the same brightness level as presented at the top of \Fig{pair}. 
For the indoor scenes, we reduced the exposure time by 1, 2, 3, 4, and 6 times.
Note that low-light and well-lit images of a pair are captured with the same exposure time.

Finally, we collected 2,556 pairs of low-light and well-lit images of 251 scenes; 2,065 pairs of 201 scenes are used for training, and the remaining 491 pairs of 50 scenes are kept for testing.
We manually annotated a bounding box and 14 body joints for each person using well-lit images following the CrowdPose dataset~\cite{Li_2019_CVPR}, and collected annotations for 14,215 human instances. 
The annotations are used as the ground-truth labels for both low-light and well-lit images as the images are spatially aligned.

While the dedicated camera system allows capturing real low-light images paired with well-lit images, it may introduce two limitations regarding the generalization ability of the pose estimation method.
First, the dataset does not cover diverse cameras as the system is designed with specific camera modules.
Second, low-light images captured with an ND filter may have different characteristics than low-light images captured at night.
Thus, for the evaluation of the generalization ability, we also propose another dataset, named ExLPose-OCN, that provides extremely low-light images captured by Other Cameras at Night.
Specifically, the ExLPose-OCN dataset provides images captured using a DSLR camera (A7M3) and a compact camera (RICHO3) and manually annotated ground-truth labels, but no well-lit images. The images are in JPEG format. For each camera, 180 images are provided.
\Tbl{dataset_statistics} summarizes statistics of the ExLPose and ExLPose-OCN datasets, and \Fig{overview_dataset} and \Fig{overview_dataset_oc} show example images of the two datasets.

%% file: Paper/Figures/Dual_camera.tex
\begin{figure}[!t]
    \centering
    \includegraphics[width=1.0\linewidth]{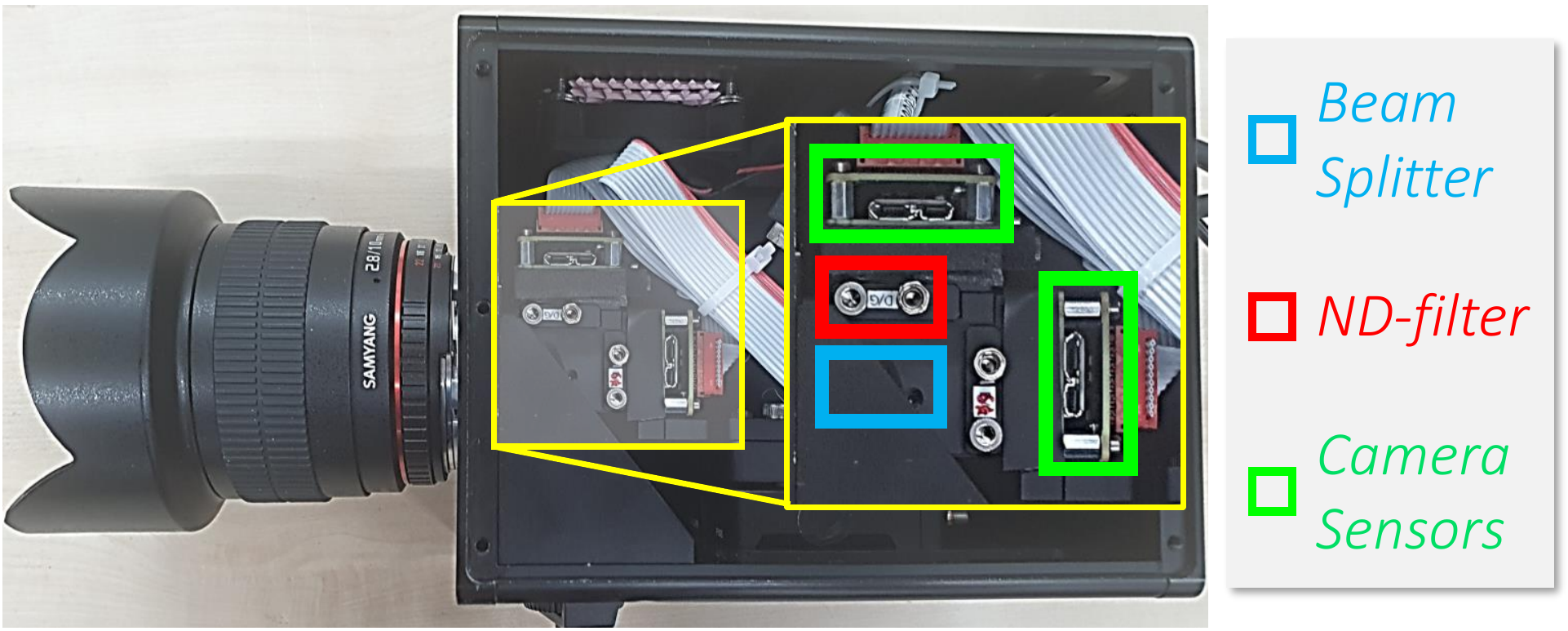} 
    \vspace{-6.0mm} 
    \caption{Our dual-camera system that consists of two camera modules with a beam splitter and an ND filter.}
    \label{fig:system}
    \vspaceBottomFigure
\end{figure}

%% file: Paper/Figures/ExLPose_pair.tex
\begin{figure}[!t]
    \centering
    \includegraphics[width=1.0\linewidth]{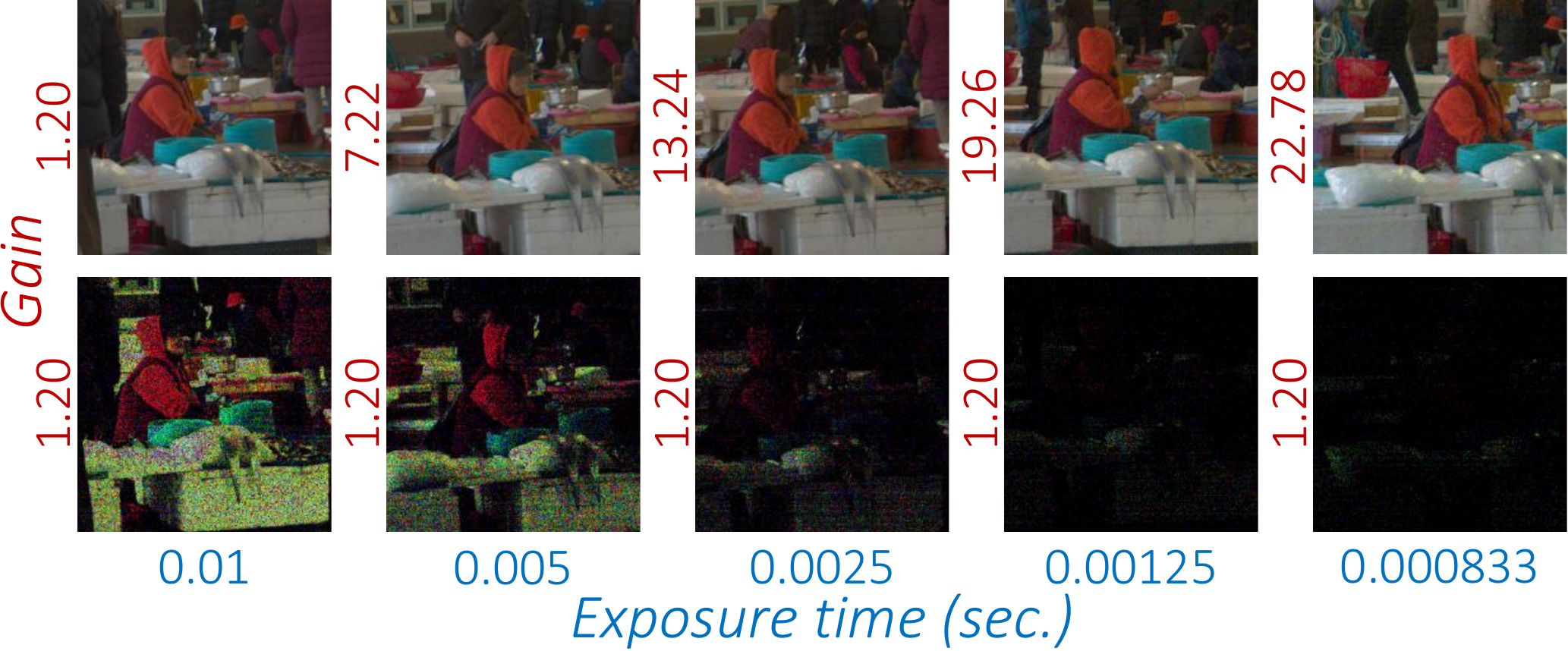}
    \vspace{-7mm}
    \caption{Pairs of well-lit (\textit{top}) and low-light (\textit{bottom}) images captured in the same scene. The low-light images are scaled by 30 times for visualization.}
    \label{fig:pair}
    \vspaceBottomFigure
\end{figure}

%% file: Paper/Tables/ExLPose_statistics.tex
\begin{table*}[!t]
\centering
\renewcommand{\arraystretch}{1.0}
\scalebox{0.83}{
\begin{tabular}{lccccccccc}
\toprule
&          & $\#$Scenes  & $\#$Images & $\#$Instances & Camera & Mean Intensity & Paired Well-lit & Resolution                       \\ \midrule
\multirow{2}{*}{ExLPose}
&Train & 201              & 2,065     & 11,405        & \multirow{2}{*}{daA1920-160uc}  & \multirow{2}{*}{2.0 (low-light) / 90.5 (well-lit)} & \multirow{2}{*}{\checkmark} &  \multirow{2}{*}{1920$\times$1200}\\
&Test  & 50              & 491       & 2,810         &                            &      &   &                     \\ \midrule
\multirow{2}{*}{ExLPose-OCN}
&\multirow{2}{*}{Test}  & \multirow{2}{*}{-}               & 180       & 466        &   A7M3  & 3.8 &  \multirow{2}{*}{}& \multirow{2}{*}{6000$\times$4000} \\ 
&                         &         &        180       & 524         &            RICOH3       &  5.6       &                          &    \\ 
\bottomrule
\end{tabular}
}
\vspace{-2mm}
\caption{Statistics of the ExLPose and ExLPose-OCN datasets.}
\vspace{-2mm}
\label{tab:dataset_statistics}
\end{table*}

%% file: Paper/Figures/ExLPose_examples.tex
\begin{figure*}[t!]

\centering
\begin{minipage}{0.62\linewidth}
\centering   
\includegraphics[width=\linewidth]{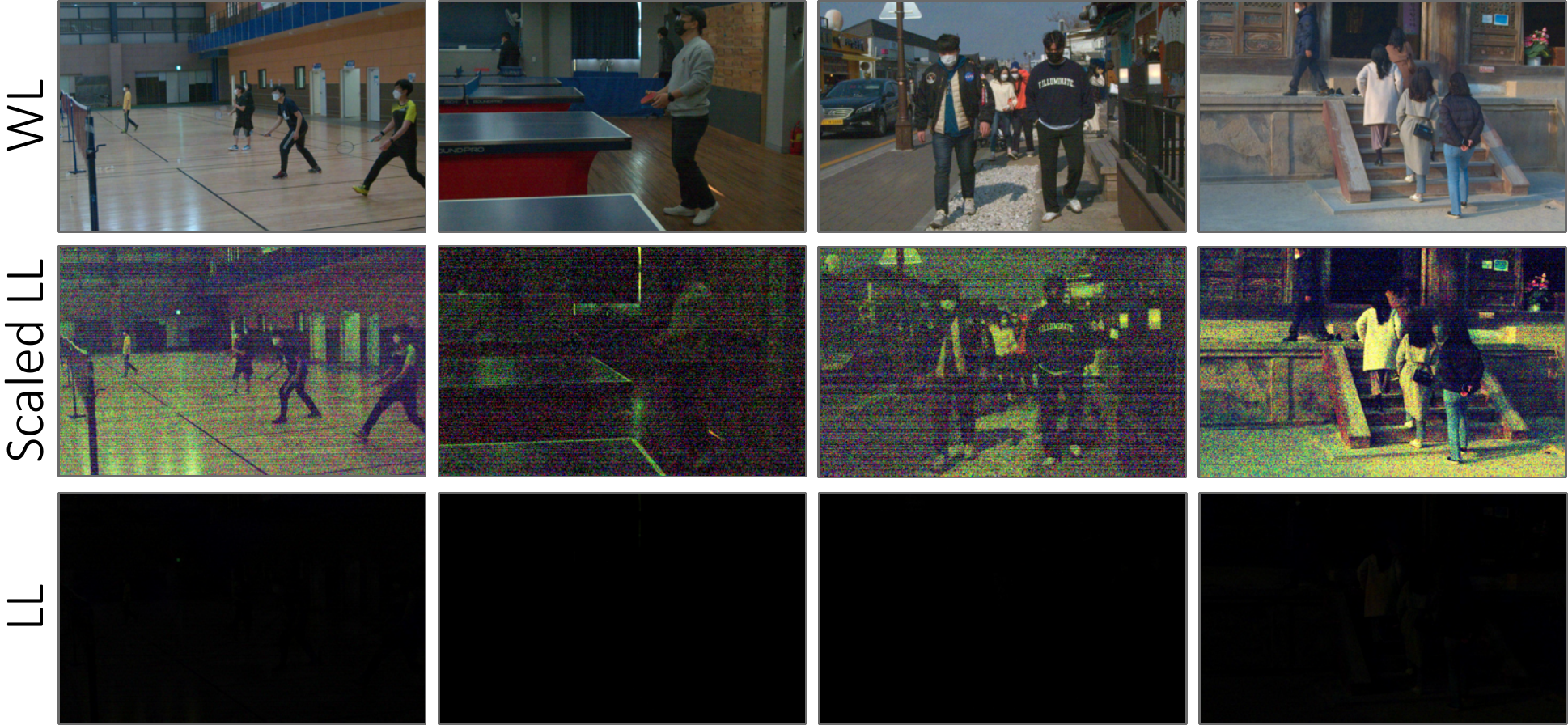} 
    \vspace{-6mm}
    \caption{
    Example images of the ExLPose dataset. WL and LL indicate well-lit and low-light images, respectively.
    Scaled LL denotes low-light images with intensities scaled up for visualization.
    } \label{fig:overview_dataset}
\end{minipage}
\hspace{4mm}
\begin{minipage}{0.33\linewidth}
\centering
    \vspace{-3mm}
    \includegraphics[width=1.0\linewidth]{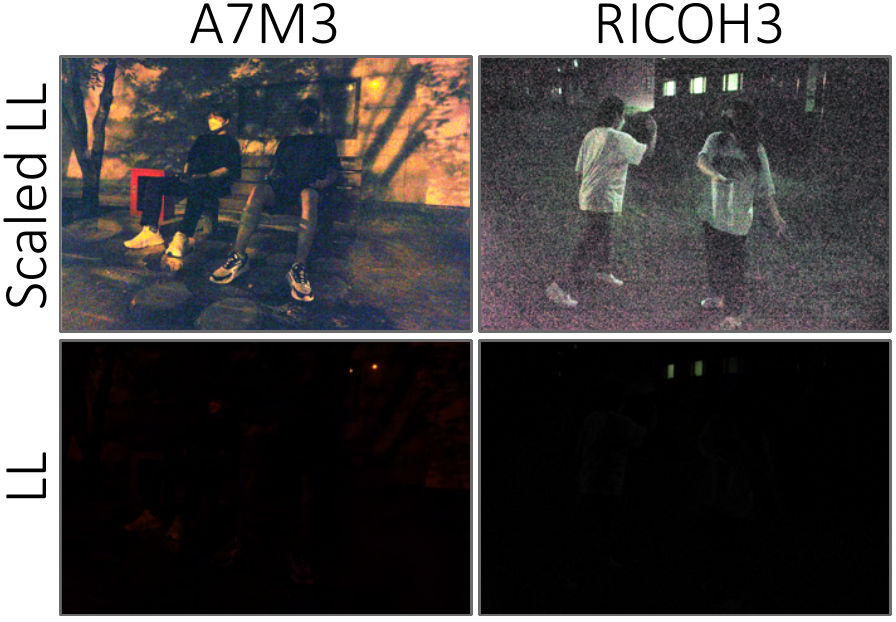}
  \vspace{-5mm}
    \caption{
    Example images of the ExLPose-OCN dataset. 
    LL and Scaled LL indicate low-light images with the original intensities and intensities scaled up for visualization, respectively.
    } \label{fig:overview_dataset_oc}
    \vspace{-2mm}
\end{minipage}
\vspaceBottomFigure
\end{figure*}

%% file: Paper/_4_method.tex
\section{Proposed Method}
\input{Paper/Figures/Architecture}

As our target model deals with extremely low-light images, it suffers from the significantly low quality of inputs.
To alleviate this, we propose a new method for learning the target model using the paired well-lit images as privileged information~\cite{LUPI_2009}, additional high-quality input data accessible only in training.
Our method introduces another model called \emph{teacher} that takes the privileged information as input and provides rich supervision to the target model called \emph{student}.
This method for learning using privileged information (LUPI) allows the student to simulate the internal behavior of the teacher as well as learn to predict human poses.

To further exploit the privileged information, we design a single concise architecture that integrates the teacher and the student.
The key idea is to let them use separate batch normalization (BN) layers while sharing all the other parameters of the network; following the domain-specific batch normalization~\cite{chang2019domain} we call such a set of separate BNs lighting-condition specific BN (LSBN).
This design choice allows the student to enjoy the strong representation learned using the well-lit images (\ie, the privileged information) while capturing specific characteristics of low-light images through the separate BN parameters. 

Our model architecture and LUPI strategy are depicted in \Fig{pipeline}.
Note that, before being fed to the student, low-light images are scaled automatically by adjusting their average pixel intensity value to a predefined constant.
On the other hand, the teacher takes as input well-lit images as-is.
Both of the teacher and the student are trained by a common pose estimation loss:
\begin{equation} 
\mathcal{L}_\textrm{pose}(\mathbf{P}, \mathbf{Y}) = \frac{1}{K}\sum_{i=1}^{K}||\mathbf{P}_i - \mathbf{Y}_i||_2^2,
\label{eq:pose}
\end{equation}
where $P_i$ and $Y_i$ denote the predicted heatmap and ground-truth heatmap for the $i$-th joint, and $K$ refers to the number of joints.
In addition to the above loss, the student takes another supervision based on the privileged information through the teacher.
The remaining part of this section elaborates on LSBN and LUPI.

\subsection{LSBN}
\label{sec:LSBN}
For each iteration of training, a low-light image $I^{\textrm{low}}$ and its well-lit counterpart $I^{\textrm{well}}$ are given together as input and processed by different BNs in LSBNs according to their light conditions.
Each LSBN layer contains two BNs, each of which has its own affine transform parameters, $(\gamma^\textrm{low}, \beta^\textrm{low})$ for low-light and $(\gamma^\textrm{well}, \beta^\textrm{well})$ for well-lit.
Within a mini-batch of $N$ samples, an LSBN layer whitens input activations and transforms them using the lighting-condition specific affine transform parameters in a channel-wise manner.
Let $\mathbf{x} \in \mathbb{R}^{N\times H\times W}$ denote a channel of activations computed from $N$ images of a specific lighting condition, and $\lambda$ be an indicator that returns 1 if the lighting condition of $\mathbf{x}$ is `low-light' and 0 otherwise.
The output of the LSBN layer taking $\mathbf{x}$ and $\lambda$ as inputs is given by 
\begin{align} 
\textrm{LSBN}(\mathbf{x}, \lambda) & = \lambda\bigg(\gamma^\textrm{low} \cdot \frac{\mathbf{x} - \mu}{\sqrt{\sigma^2+\epsilon}} + \beta^\textrm{low}\bigg) \nonumber \\
& + (1-\lambda)\bigg(\gamma^\textrm{well} \cdot \frac{\mathbf{x} - \mu}{\sqrt{\sigma^2+\epsilon}} + \beta^\textrm{well}\bigg),
\label{eq:lsbn}
\end{align} 
where $\mu$ and $\sigma^2$ denote mean and variance of the activations in $\mathbf{x}$, respectively, and $\epsilon$ is a small constant adopted for numerical stability.

\subsection{LUPI}
\label{sec:LUPI}

Since low-light and well-lit images of a pair share the same content in our dataset, we argue that the gap between their predictions will be largely affected by their style difference.\footnote{It has been known that an image is separated into content and style~\cite{gatys2016image, kotovenko2019content}. Since a pair of low-light and well-lit images in ExLPose have the same content due to our camera system, they are different only in style.} 
In our case, the style of an image is determined by high-frequency noise patterns and its overall intensity.
Hence, in our LUPI strategy, the teacher provides neural styles of its intermediate feature maps as additional supervision to the student so that the student learns to fill the style gap between low-light and well-lit images in feature spaces; this approach eventually leads to a learned representation insensitive to varying lighting conditions.

To implement the above idea, we adopt as a neural style representation the Gram matrix~\cite{gatys2016image}, denoted by $\textbf{G} \in \mathbb{R}^{C \times C}$, that captures correlations between $C$ channels of a feature map $\textbf{F}$.  
Specifically, it is computed by $\textbf{G}_{i,j} = \textbf{f}_{i}^\top \textbf{f}_{j}$ where $\textbf{f}_{i}$ is the vector form of the $i^\textrm{th}$ channel of $\textbf{F}$.
Let $\textbf{G}^{\textrm{low},l}$ denote the neural style of a low-light image computed from the feature map of the $l^\textrm{th}$ layer of the student, and similarly, $\textbf{G}^{\textrm{well},l}$ be the neural style of the coupled well-lit image computed from the feature map of the $l^\textrm{th}$ layer of the teacher.
Then our LUPI strategy minimizes the following loss with respect to $\textbf{G}^{\textrm{low},l}$ of all predefined layers so that the neural style of the student approximates that of the teacher:
\begin{equation} 
\mathcal{L}_\textrm{LUPI} = \sum_{l} {\frac{1}{4 C_l^2 N_l^2} \sum_{i=1}^{C_l}\sum_{j=1}^{C_l} \Big(  \textbf{G}_{i,j}^{\textrm{low},l} - \textbf{G}_{i,j}^{\textrm{well},l} \Big)^2},
\label{eq:lupi_loss}
\end{equation}
where $C_l$ and $N_l$ are the number of channels and the spatial size of the $l^\textrm{th}$ feature map, respectively.

\subsection{Empirical Justification}
To investigate the impact of LUPI, we first demonstrate that it reduces the style gaps between different lighting conditions.
To this end, we compute the average Hausdorff distance~\cite{dubuisson1994modified} between the sets of Gram matrices of different lighting conditions before and after applying LUPI. 
\Fig{Empirical_analysis}(a) shows that the style gaps between different lighting conditions are effectively reduced by LUPI as intended.

It is also empirically examined if LSBN and LUPI of our method eventually lead to a model insensitive to lighting conditions. 
For this purpose, each lighting condition is represented by the set of features computed from associated images, and discrepancies between different lighting conditions are estimated by the average Hausdorff distances between such sets.
As shown in \Fig{Empirical_analysis}(b), LSBN matches the feature distributions of different lighting conditions, compared to the
model trained with only pose estimation loss for both low-light and well-lit images, denoted by Baseline-all.
LUPI even further closes the gaps, leading to the representation 
that well aligns two different lighting conditions.
\input{Paper/Figures/Empirical_analysis}

%% file: Paper/Figures/Architecture.tex
\begin{figure*}[t]
    \centering
    \vspace{0mm}
    \includegraphics[width=0.99\linewidth]{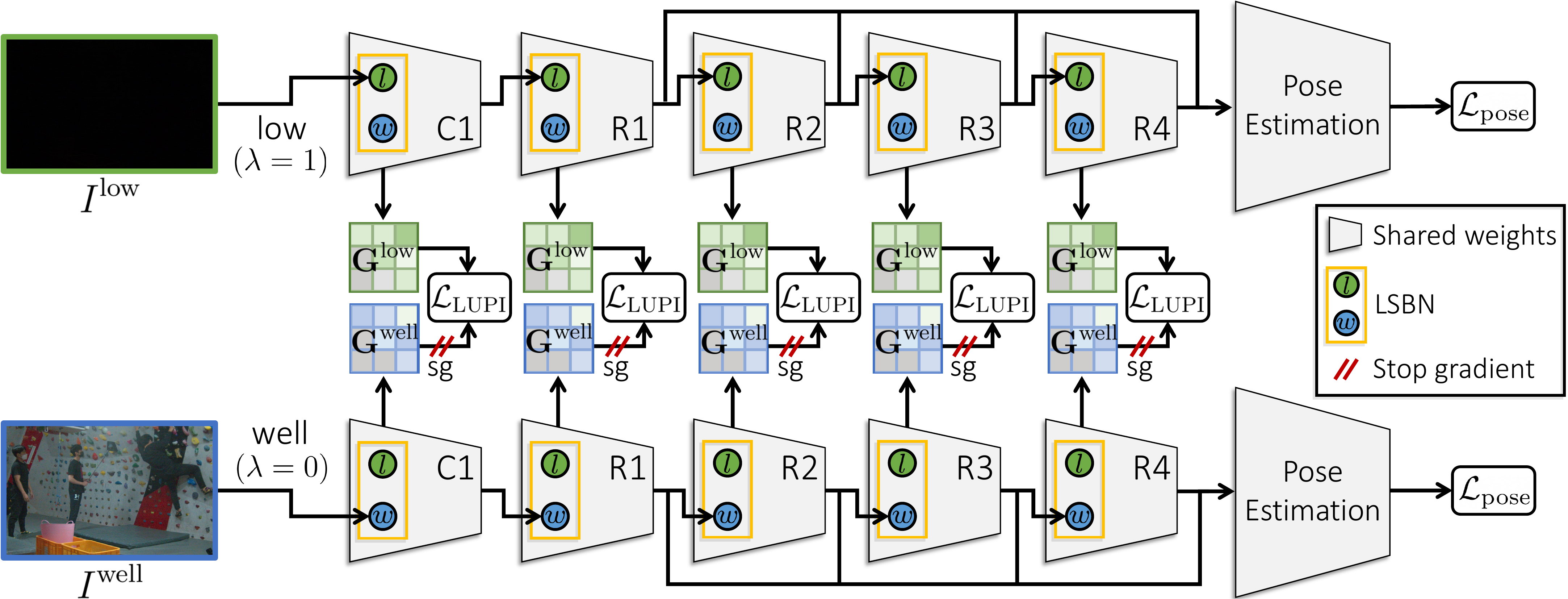}
    \vspace{-2mm}
\caption{
The proposed model architecture and training strategy. 
Both of teacher (\textit{bottom}) and student (\textit{top}) are trained by the same pose estimation loss, and student takes additional supervision from teacher through LUPI.
The loss for LUPI is applied to the feature maps of the first convolutional layer (\ie, C1) and the following four residual blocks (\ie, R1--R4) of a ResNet backbone.
Teacher and student share all the parameters except LSBNs.
Details of the pose estimation module are presented in Fig.~\ref{fig:pose_architecture} of the supplementary material.
}
\label{fig:pipeline}
\vspaceBottomFigure
\end{figure*}

%% file: Paper/Figures/Empirical_analysis.tex
\begin{figure}[t]
  \centering
    \includegraphics[width=\linewidth]{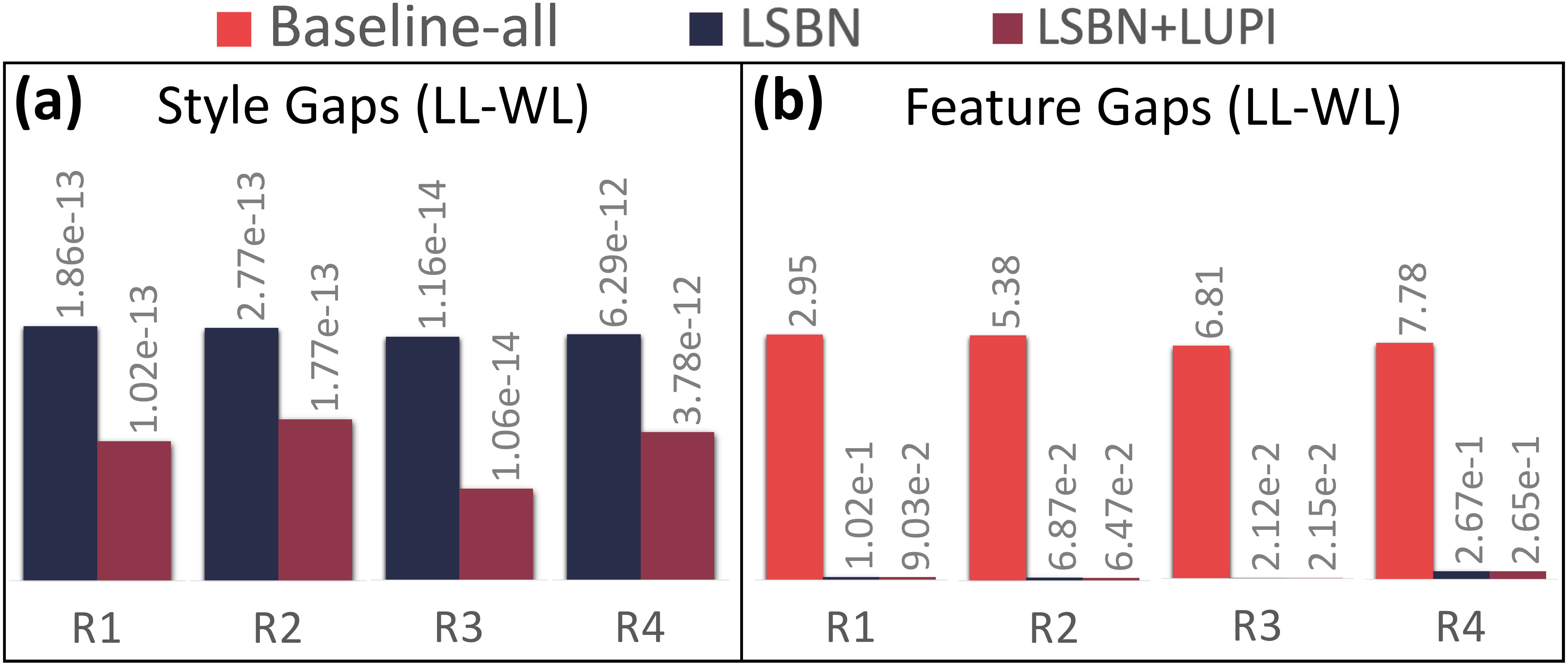}
    \vspace{-6mm}
    \caption{
      Quantitative analysis on (a) style gaps and (b) feature gaps measured by the average Hausdorff distance~\cite{dubuisson1994modified} between low-light and well-lit conditions.
      The gaps are measured on each level of the four residual blocks (\ie, R1-R4) of the ResNet backbone while considering each lighting condition as a set of styles of associated images in (a) and their features in (b).
      } \label{fig:Empirical_analysis}
\vspaceBottomFigure
\end{figure}

%% file: Paper/_5_experiments.tex
\vspace{-3mm}
\section{Experiments}
\subsection{Single-person Pose Estimation}\label{sec:single_pose}

To ease the difficulty of the problem and focus solely on pose estimation, we first tackle single-person pose estimation with the assumption that ground-truth bounding boxes are given for individuals.

\noindent\textbf{Implementation Details.}
We adopt Cascaded Pyramid Network (CPN)~\cite{chen2018cascaded} with ResNet-50~\cite{resnet} backbone as our pose estimation network, which is pre-trained on the ImageNet dataset~\cite{Imagenet}.
All BNs in the backbone are replaced with LSBNs, and LUPI is applied to the outputs of the first convolution layer and the following four residual blocks.
The average channel intensity of an input low-light image is automatically adjusted to 0.4 before being fed to the student network following Zheng~\etal~\cite{Zheng_2020_CVPR}.
More details are given in Sec.~\ref{appendix:details} of the supplement.

\noindent\textbf{Evaluation Protocol.}
We report the standard average precision (AP) scores based on object keypoint similarity following CrowdPose~\cite{li2019crowdpose}.
Our method and competitors are evaluated on both low-light (LL) and well-lit (WL) images.
Low-light images for testing are further divided into three subsets according to their relative difficulty, low-light normal (LL-N), low-light hard (LL-H), and low-light extreme (LL-E), by the gain values of the coupled well-lit images.
Specifically, they are split into 167 LL-N images, 169 LL-H images, and 155 LL-E images by applying two thresholds, 15 and 24, to their gain values.
The mean pixel intensities of LL-N, LL-H, and LL-E images are 3.2, 1.4, and 0.9, respectively.
Note that images of all three subsets are captured in extremely low-light conditions and hard to be recognized by humans.
The union of the three low-light test splits is denoted as low-light all (LL-A).

\input{Paper/Tables/Main_table}
\input{Paper/Figures/Qualitative_results}

\vspace{-2mm}
\subsubsection{Quantitative Results on the ExLPose Dataset}
\label{sec:quan_analysis_PID}

Our method is compared with potential solutions to the target task on the five test splits of the ExLPose dataset, \ie, LL-N, LL-H, LL-E, LL-A, and WL. 
The solutions include baselines that train the pose estimation model directly with the ExLPose dataset, 
those incorporating low-light image enhancement techniques as pre-processing, and domain adaptation methods that consider different lighting conditions as different domains.
Specifically, the baselines are CPNs trained on low-light images, well-lit images, or both of them using the pose estimation loss only, which are denoted by Baseline-low, Baseline-well, and Baseline-all, respectively.
Also, when a low-light image enhancement technique is incorporated, low-light training images are first enhanced and both of the enhanced images and well-lit images are used to train CPN, for which low-light test images are also enhanced by the same technique; we found that this strategy maximizes the advantage of image enhancement as shown in Table.~\ref{tab:variant_enhancement} of the supplement. 
Pose estimation performance of ours and these solutions is given in \Tbl{main_table}. 

Our method clearly outperforms all the three baselines in the four low-light splits, and is even on par with Baseline-well in the well-lit split. 
As expected, Baseline-low and Baseline-well are significantly inferior in the well-lit and low-light splits, respectively, 
due to the substantial gap between training and testing images in lighting conditions.
Baseline-all achieves the best among the baselines, but its performance is still limited compared with other approaches including ours.
These results suggest that it is not straightforward to learn a common model working under both low-light and well-lit conditions, as also reported in~\cite{loh2019getting,morawski2021nod}.
In contrast to these early findings, we successfully manage to train a single model that performs well under both of the two lighting conditions through LSBN and LUPI.

To evaluate the efficacy of low-light image enhancement, we adopt two enhancement methods: LLFlow~\cite{wang2021low} as a learning-based method and LIME~\cite{LIME_2017_TIP} as a traditional method based on the Retinex theory. 
The combinations of these enhancement techniques and CPN are denoted as `LLFlow + Baseline-all' and `LIME + Baseline-all'.
As shown in the table, low-light image enhancement helps improve performance in two evaluation settings, LL-N and LL-H, but it rather degrades performance in LL-E. 
Furthermore, it additionally imposes immense inference latency, and demands a large amount of memory footprint when adopting learning-based methods like LLFlow.
On the other hand, our method outperforms them by large margins in all the five splits with only a small number of additional parameters and no additional inference latency.

Finally, our method is compared with two domain adaptation (DA) methods: DANN~\cite{dann} for feature-level DA and AdvEnt~\cite{vu2019advent} for output-level DA. 
Note that existing DA methods for 2D human pose estimation~\cite{jakab2020self,cao2019cross} are not compared since they are essentially based on DANN~\cite{dann}.
For training CPN with these methods, we assign well-lit images to a source domain and low-light images to a target domain, and utilize pose labels of both domains.
The results in the table show that the direct adaptation between low-light and well-lit conditions is not effective due to the large domain gap, which suggests the necessity of LSBN. 
Thanks to LSBN and LUPI, our method clearly surpasses the two methods in every split.
Further analysis on the impact of LSBN and LUPI is presented in Sec.~\ref{appendix:ablation} of the supplement.

\vspace{-2mm}
\subsubsection{Qualitative Results on the ExLPose Dataset}
\label{sec:qual_analysis_PID}

Our method is qualitatively compared with Baseline-all, DANN, LIME + Baseline-all in \Fig{qualitative_results}.
As shown in the figure, 
Baseline-all and DANN fail to predict poses frequently.
LIME + Baseline-all performs best among the competitors, but often fails to capture details of poses,
in particular under more difficult low-light conditions.
Our method clearly exhibits the best results; it estimates human poses accurately even under the \mbox{LL-E} condition with severe noises. 

\input{Paper/Tables/ExLPose_significance}

\vspace{-2mm}
\subsubsection{Significance of the ExLPose Dataset}

The ExLPose dataset has two significant properties: It provides pairs of aligned low-light and well-lit images, and it thus enables accurate labeling of low-light images. 
The impact of these properties is investigated by additional experiments, whose results are summarized in Table~\ref{tab:ablation}.

First, we study the importance of pairing low-light and well-lit images. 
To this end, we compare our method (LSBN + LUPI w/pair) with its variant (LSBN+LUPI w/o pair) disregarding the pair relations, trained with unrelated low-light and well-lit images.
As demonstrated in the table, our method clearly outperforms the variant, which suggests the contribution of pairing low-light and well-lit images.
Next, we investigate the impact of accurate pose labels for low-light images by comparing our method with its another variant trained in an unsupervised domain adaptation (UDA) setting.
To be specific, the second variant (Ours-UDA w/o pair) is trained with both LSBN and LUPI using labeled well-lit images and unlabeled low-light images, which are unpaired, following the common problem setting of UDA.
As demonstrated in the table, this variant performs worst, which justifies the significance of pose labeling on low-light images as well as that of pairing low-light and well-lit images in our dataset.

\input{Paper/Tables/Loss_ablation}

\vspace{-2mm}
\subsubsection{Ablation Study}\label{sec:ablation_study}
We investigate the impact of LSBN and LUPI by the ablation study in \Tbl{ablation_main}.
In the table, LSBN and LUPI denote variants of ours trained with either LSBN or LUPI, respectively.
The inferior performance of LUPI implies that LSBN is essential to bridge the large gap between low-light and well-lit conditions effectively. 
For the same reason, LSBN significantly improves performance over Baseline-all. 
LUPI further contributes to the outstanding performance of our method: It improves LSBN substantially in every split when integrated.
For an in-depth analysis on the impact of LUPI, our method is also compared with another variant (LSBN + LUPI-\textit{feat}) that directly approximates features of the teacher instead of its neural styles.
The performance gap between our model and this variant empirically justifies the use of neural styles for LUPI.

\input{Paper/Tables/ExLPose_OCN}

\subsubsection{Results on the ExLPose-OCN Dataset}
\label{sec:eval_PID-OCN}

To demonstrate the generalization capability of our method,
we evaluate our method and the other solutions on the ExLPose-OCN dataset. 
As shown in \Tbl{camera_table}, all enhancement and DA methods do not perform well on the ExLPose-OCN dataset; their performance is even inferior to that of Baseline-all (Base-all).
In contrast, our method achieves the best on the ExLPose-OCN dataset also, which suggests that it improves the generalization ability as well as pose estimation accuracy.
\Fig{ExLPose_ocn_qual} also demonstrates that our method qualitatively outperforms Baseline-all (Base-all).
Additional qualitative results are given in Sec.~\ref{appendix:qualatitive_ExLPose-OCN} of the supplement.

\input{Paper/Tables/Multipose}

\vspace{+1.0mm}
\subsection{Multi-person Pose Estimation}
\label{sec:multi-person}

Our method and most of the potential solutions can be extended to multi-person pose estimation since they are applicable to person detection as well as pose estimation. 
To this end, we train Cascade R-CNN~\cite{cai2018cascade} for person detection with the ExLPose dataset, and then utilize bounding boxes predicted by the detector instead of ground-truths; 
technical details and results of person detection in low-light conditions are presented in Sec.~\ref{appendix:networks} and Sec.~\ref{appendix:detection} of the supplement.
In \Tbl{multi-person}, each method is applied to both person detection and pose estimation models, and our method outperforms all the others in every split. 
As shown in the bottom row of Fig.~\ref{fig:qualitative_results}, our method qualitatively outperforms other solutions. 

%% file: Paper/Tables/Main_table.tex
\begin{table*}[!t]
\centering
\renewcommand{\arraystretch}{1.0}
\scalebox{0.97}{\begin{tabular}{lcccccccccc}
\toprule
  & \multicolumn{3}{c}{Training data}  & \multicolumn{5}{c}{AP@0.5:0.95}         &   \multirow{2}{*}{Param. (M)}    & \multirow{2}{*}{Latency (sec)} \\ 
    \cline{5-9}
  & LL & WL & Enhanced-LL & LL-N     &  LL-H       &  LL-E   & LL-A  &  WL     &    &      \\ \midrule
Baseline-low  & \checkmark & &   &    32.6      &     25.1      &    13.8     &  24.6    & ~~1.6    &  27.37  &  1.07   \\
Baseline-well & & \checkmark &    &    23.5        &    ~~7.5       &    ~~1.1     &  11.5    & \textbf{68.8}    &  27.37  &  1.07  \\
Baseline-all & \checkmark & \checkmark &   &   33.8        &   	25.4        & 	\underline{14.3}	 &  25.4       &   57.9        &  27.37  &  1.07           \\\midrule
LLFlow + Baseline-all & & \checkmark & \checkmark  & 35.2   &    	20.1	    &	~~8.3	   & 22.1  & 	65.1    &  66.23  & 3.34 \\
LIME + Baseline-all  & & \checkmark & \checkmark  & \underline{38.3}	& \underline{25.6}	 &	12.5 & \underline{26.6}	  & 63.0   &  27.37  &  1.65  \\\midrule
DANN   & \checkmark & \checkmark &    & 34.9	   & 	24.9	 & 	13.3		 & 25.4   &	58.6    &  27.37  &  1.07	       \\
AdvEnt   & \checkmark & \checkmark &  & 	35.6  & 	23.5	 &	8.8  &	23.8   &	62.4   &  27.37  &  1.07 	  \\\midrule
Ours  & \checkmark & \checkmark &   & \textbf{42.3}   & \textbf{34.0}   & \textbf{18.6}       &  \textbf{32.7}  &	\underline{68.5}      &  27.53  &  1.07    \\ \bottomrule
\end{tabular}}
\vspace{-2mm} 
\caption{
Pose estimation accuracy in AP@0.5:0.95 on the ExLPose dataset. In the case of training data, LL, WL, and Enhanced-LL indicate low-light, well-lit, and enhanced low-light images, respectively. In the case of evaluation splits, LL-E, LL-N, LL-H, LL-A, WL stand for low-light-easy, Low-light-normal, low-light-hard, low-light-all, and well-lit splits, respectively.  
The number of parameters and prediction latency of each method are reported along with the accuracy.
Baseline-low, Baseline-well, and Baseline-all are the base pose estimation models (\ie, CPN~\cite{chen2018cascaded}) trained on low-light, well-lit, and both, respectively.}
\label{tab:main_table}
\end{table*}

%% file: Paper/Figures/Qualitative_results.tex
\begin{figure*}[t]
    \centering
    \vspace{-2mm}
    \includegraphics[width=\linewidth]{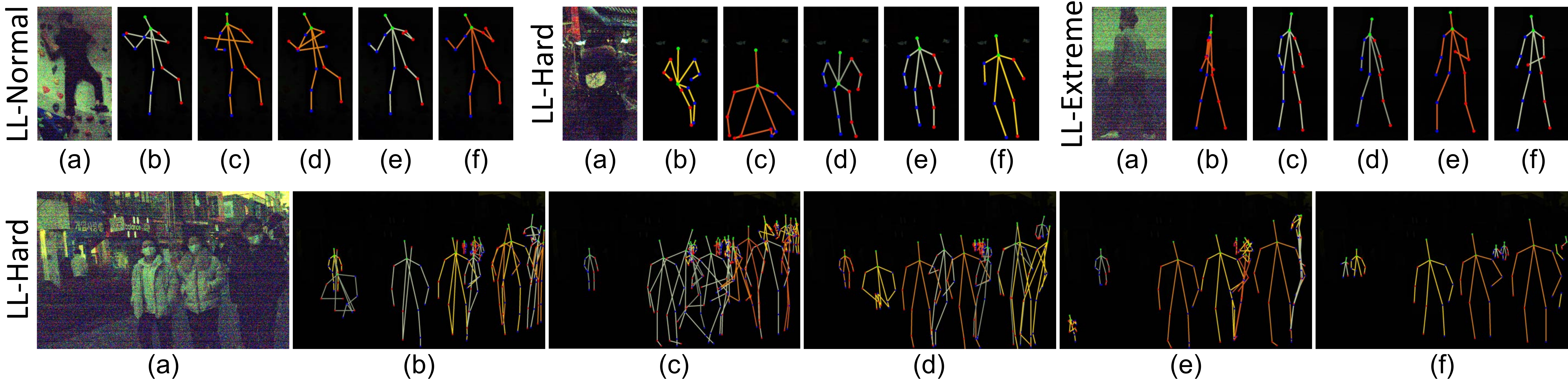}
    \vspace{-6.5mm}
\caption{
Qualitative results of single-person (\textit{top}) and multi-person (\textit{bottom}) pose estimation on the ExLPose dataset.
Predicted poses and labels are visualized on corresponding low-light images.
(a) Scaled low-light images. (b) Baseline-all. (c) DANN. (d) LIME + Baseline-all. (e) Ours. (f) Ground-truth. More results can be found in Sec.~\ref{appendix:additional_qual} of the supplementary material.
} \label{fig:qualitative_results}
\vspaceBottomFigure
\end{figure*}

%% file: Paper/Tables/ExLPose_significance.tex
\begin{table}[!t]
    \centering
    \renewcommand{\arraystretch}{1.0}
    \scalebox{0.83}{\begin{tabular}{cccc|ccccc}
    \toprule
     &\multicolumn{2}{c}{Label}   & Paired   &    \multirow{2}{*}{LL-N} & \multirow{2}{*}{LL-H}  & \multirow{2}{*}{LL-E} &   \multirow{2}{*}{LL-A} & \multirow{2}{*}{WL}   \\ 
     \cline{2-3}
     & WL    &  LL    &  WL-LL  &    &      &   &   &   \\ \hline
     (a) & \checkmark    & \checkmark    & \checkmark      & \textbf{42.3}	   & \textbf{34.0}	 & \textbf{18.5} &	\textbf{32.7} & \underline{68.5} 	       \\
     (b) &\checkmark    & \checkmark     &  & \underline{41.1}	   & \underline{30.7}	 & \underline{15.5} &	\underline{30.2} &  65.9  	       \\
     (c) &\checkmark  &     &   &   5.0    &     4.7      &  4.5   & 4.6 &   52.8      \\
     \bottomrule
    \end{tabular}}
    \vspace{-2mm}
    \caption{Analysis on the significance of the paired low-light and well-lit images of the ExLPose dataset.
    (a) LSBN + LUPI w/pair (Ours) (b) LSBN + LUPI w/o pair. (c) Ours-UDA w/o pair.
    }
    \label{tab:ablation}
    \vspaceBottomTable
\end{table}

%% file: Paper/Tables/Loss_ablation.tex
\begin{table}[t]
    \centering
    \centering
    \renewcommand{\arraystretch}{1.0}
    \scalebox{0.88}{\begin{tabular}{lccccc}
        \toprule
        AP@0.5:0.95     & LL-N     &  LL-H       &  LL-E   & LL-A  &  WL \\\midrule
        Baseline-all   &   33.8        &   	25.4        & 	14.3	 &  25.4       &   57.9   \\ \midrule
        LUPI & 	34.2  & 	23.1	 &	11.2  &	24.0   &	61.7  	  \\ 
        LSBN &  39.0	 &     30.2	 &   \underline{18.3}	  &     30.1	 &  \underline{67.2}       \\\midrule
        LSBN + LUPI--\emph{feat}     &  \underline{39.7}	&    \underline{30.8}	&  17.4    &   	\underline{30.4}  &	65.2  	    \\
        LSBN + LUPI (Ours) & \textbf{42.3}   & \textbf{34.0}   & \textbf{18.5}       &  \textbf{32.7}  &	\textbf{68.5}     \\ \bottomrule
    \end{tabular}}
    \vspace{-2mm}
    \caption{Analysis on the impact of LUPI and LSBN.}\label{tab:ablation_main}
    \vspace{-1.0mm}
\end{table}

%% file: Paper/Tables/ExLPose_OCN.tex
\begin{table}[t!]
\vspace{-2mm}
  \centering
  \hspace{-1mm}
  \scalebox{0.94}{\begin{minipage}{0.59\linewidth}
        \centering
      \vspace{0mm}
      \hspace{-2mm}
      \normalsize
      \renewcommand{\arraystretch}{1.05}
        \scalebox{0.7}{
        \begin{tabular}{lcccc}
        \toprule
        AP@0.5:0.95     &  A7M3  &  RICOH3  &   Avg.    \\ \midrule
        Base-low    &  23.7	 &    23.9	& 23.8 \\
        Base-well   &  15.2	 &    15.6  & 15.4 \\
        Base-all    &  32.8  &   \underline{31.7}   & \underline{32.2}    \\\midrule
        LLFlow + Base-all  &  25.6  & 28.2 	  & 27.0 \\
        LIME + Base-all   & \underline{33.2} & 28.4	  & 30.7 \\\midrule
        DANN     & 	27.9 &  30.6	 & 29.3  \\
        AdvEnt      & 28.2 & 29.0    & 28.6 \\\midrule
        Ours  &  	\textbf{35.3}   & \textbf{35.1} &  \textbf{35.2} \\ \bottomrule
        \end{tabular}}
        \vspace{-2mm}
        \caption{Quantitative results on the ExLPose-OCN dataset. Base denotes Baseline (\ie, CPN~\cite{chen2018cascaded}).}
        \label{tab:camera_table}
  \end{minipage}}%
  \quad 
  \scalebox{0.94}{\begin{minipage}{.42\linewidth}
    \centering
    \scalebox{1.01}{\includegraphics[width=\linewidth]{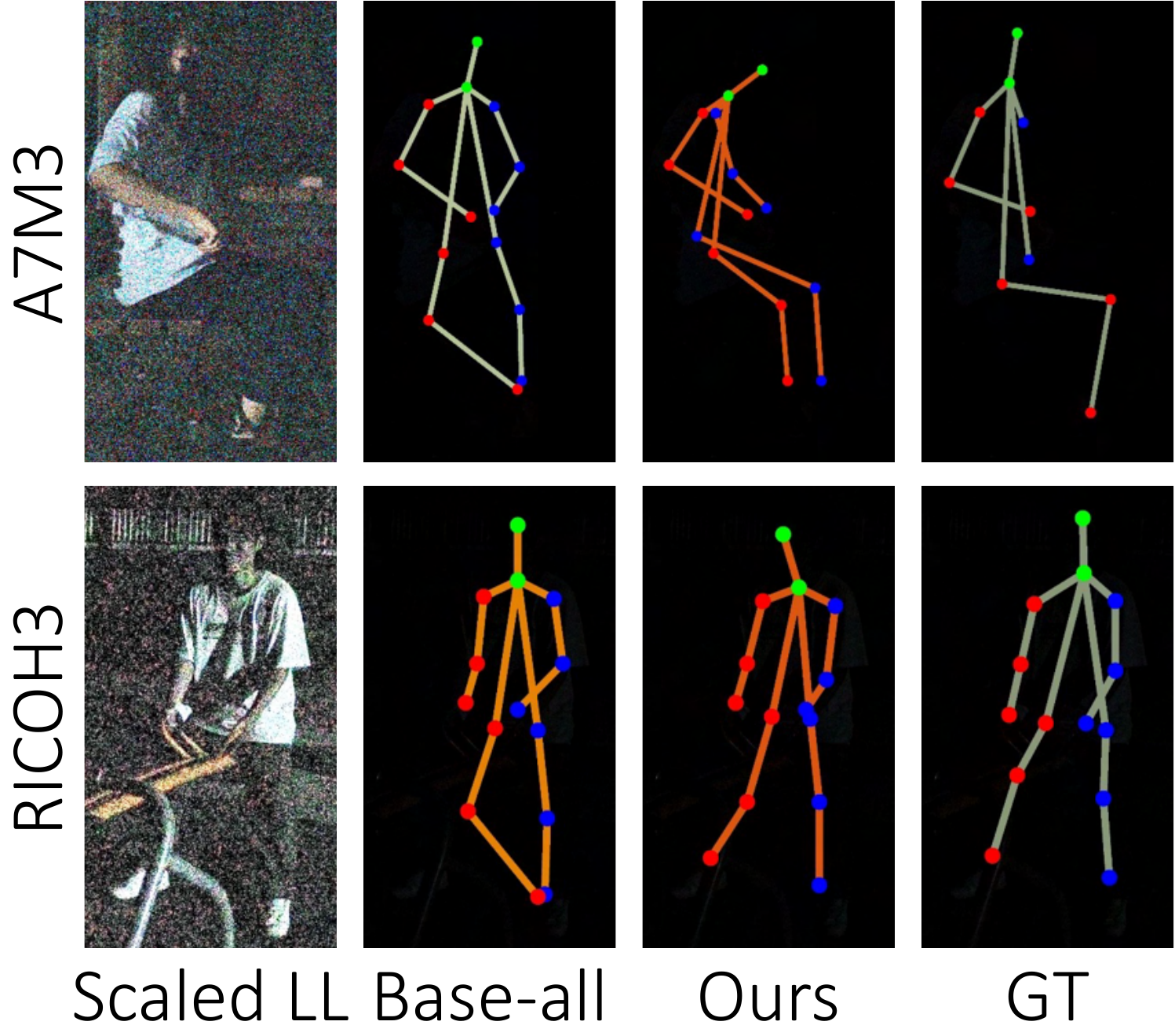}}
    \vspace{-6mm}
    \captionof{figure}{Qualitative results on the ExLPose-OCN dataset. Results are visualized on low-light images.
    }
    \label{fig:ExLPose_ocn_qual}
    \vspace{0mm}
  \end{minipage}} 
  \vspace{-4mm}

\end{table}

%% file: Paper/Tables/Multipose.tex
\begin{table}[t]
    \centering
    \scalebox{0.87}{\begin{tabular}{lccccc}
        \toprule
        AP@0.5:0.95        & LL-N         &  LL-H       &  LL-E   &  LL-A     &  WL     \\ \midrule
        Baseline-low         & 25.4 & 18.2 & 6.0 & 17.2 & 1.2  \\
        Baseline-well       & 15.3 & 2.7 & 0.4 & 6.7 & 59.9  \\
        Baseline-all         & 26.4 & 18.2 & 6.1 & 17.6 & 52.3  \\ \midrule
        LLFlow + Baseline-all   & 28.7 & 15.7 & 5.3 & 17.4 & \underline{60.7} \\
        LIME + Baseline-all  & \underline{31.9} & \underline{21.2} & \underline{7.6} & \underline{21.1} & 57.7    \\\midrule  
        DANN       & 28.0 & 17.5 & 5.3 & 17.8 & 52.0   \\     \midrule        
        Ours       & \textbf{35.6} & \textbf{25.0} & \textbf{11.6} & \textbf{25.0} & \textbf{61.5} 	 	       \\ \bottomrule
    \end{tabular}}
    \vspace{-2mm}
    \caption{Quantitative results of multi-person pose estimation.}\label{tab:multi-person}
    \vspaceBottomTable
    \vspace{-1mm}
\end{table}

%% file: Paper/_6_conclusion.tex
\vspace{+1.0mm}
\section{Conclusion}

We have introduced the first attempt to estimate human poses in extremely low-light images. 
To this end, we have first presented the ExLPose dataset that provides paired low-light and well-lit images with accurate human pose labels.
The novel model and training strategy also have been proposed for learning representations insensitive to lighting conditions using the paired well-lit images as privileged information.
Our method has been evaluated on real images taken under various low-light conditions, where it clearly outperforms domain adaptation and image enhancement methods.

\vspace{3mm}
{\small
\noindent \textbf{Acknowledgement.} 
This work was supported by Samsung Research Funding \& Incubation Center of Samsung Electronics under Project Number SRFC-IT1801-52.
}

%% file: Supplement/Supp.tex
\newcommand{\expnum}[2]{{#1}\mathrm{e}{-#2}}

\setcounter{section}{0}
\def\thesection{\Alph{section}}
\section*{\Large{Appendix}}
\renewcommand\thefigure{A\arabic{figure}}
\renewcommand{\thetable}{A\arabic{table}}
\setcounter{figure}{0}
\setcounter{table}{0}

This supplementary material presents additional experimental details and results that are omitted from the main paper due to the space limit.
Section~\ref{appendix:low-light_imaging} explains extremely low-light images of the ExLPose dataset.
Section~\ref{appendix:dataset_scale} shows a comparison of dataset scale with other human pose datasets.
Section~\ref{appendix:geometric_alignment} presents a detail of the geometric alignment of our dual-camera system. Section~\ref{appendix:details} describes experimental details about implementation (Sec.~\ref{appendix:implementation}) and network architectures (Sec.~\ref{appendix:networks}).
Section~\ref{appendix:empirical_analysis} shows an in-depth analysis of our method, including lighting condition analysis (Sec.~\ref{appendix:lighting_conditions}), lighting condition insensitive features analysis (Sec.~\ref{appendix:qualitative_analysis}), and additional analysis for our method (Sec.~\ref{appendix:quantitative_analysis}).
Section~\ref{appendix:Variant_enhancemnt} shows detailed results of the various combinations of the existing enhancement methods and the pose estimation network.
Section~\ref{appendix:ablation} gives extensive experimental results to investigate the components of our method, such as LSBN (Sec.~\ref{appendix:effect_LSBN}), LUPI (Sec.~\ref{appendix:feature_regression}, \ref{appendix:layer_selection}, \ref{appendix:gradient}), and intensity scaling (Sec.~\ref{appendix:scaling}).
Section~\ref{appendix:detection} presents experimental results of person detection in extremely low-light conditions. 
Finally, Section~\ref{appendix:additional_qual} shows additional qualitative results.

\section{Explanation on Extremely Low-light Images}\label{appendix:low-light_imaging}

Low-light conditions are important as they are prevalent in many scenarios with limited illumination, such as nighttime and low-light indoor environments. 
Capturing images in such conditions is challenging due to the requirement of a long exposure time, which can cause motion blur, a thorny problem to solve.
To avoid blur, a common practice is to utilize extremely low-light images captured with a short exposure time for various tasks in low-light environments, such as image enhancement~\cite{chen2018learning, Jiang_2019_ICCV, wei2020physics, wei2021physics}, action recognition~\cite{xu2020arid}, image matching~\cite{Song_2021_ICCV}, optical flow estimation~\cite{Zheng_2020_CVPR}.
In the case of human pose estimation, capturing images with a long exposure time is impossible due to the movement of humans.  
Thus, we study pose estimation under extremely low-light conditions using an extremely low-light image rather than images captured with long exposure.
The low-light images can be amplified to improve visibility using global scaling, but noise is also amplified, resulting in extremely noisy images.
\Fig{SID_and_ExLpose} shows low-light and scaled images in the SID~\cite{chen2018learning} and ExLPose datasets. The scaled images are extremely noisy since the scaling amplifies noise; the degree of noise depends on that of darkness.

\input{Supplement/Figures/SID_ExLPose}

\section{Scale Comparison to Other Datasets}\label{appendix:dataset_scale}
We compare our dataset to existing human pose estimation datasets.
As shown in the \Tbl{statistics_datasets}, the scale of our dataset is large enough since it is comparable to common datasets in terms of the number of annotated people as shown in the table.
Note that the unit of supervision in pose estimation is a human instance, not an image.

\input{Supplement/Tables/Dataset_scale}

\section{Geometric Alignment of Our Camera System}\label{appendix:geometric_alignment}

We physically align two camera modules of the dedicated dual-camera system as much as possible to capture images that are aligned with each other.
Nevertheless, there may exist a small amount of geometric misalignment between the cameras as discussed in \cite{jsrim-ECCV2020}.
Moreover, while moving around the camera system collecting the dataset, the movement of the camera system may introduce additional geometric misalignment.

To resolve this, we captured a reference image pair of a static scene before collecting data every time we moved the camera system, and estimated a homography between them.
We set the exposure time of the low-light camera module 100 times longer so that pairs of reference images have the same brightness for accurate homography estimation.
Then, we estimated a homography matrix between them using the method of~\cite{Evangelidis-TPAMI08} and aligned the collected well-lit images using the estimated homography matrix.
\Fig{geometric_alignment}(a)-(b) visualize the effect of geometric alignment using stereo-anaglyph images where a pair of well-lit and scaled low-light images from the camera modules are visualized in red and cyan.
As the figure shows, even before the geometric alignment, images from our camera system have only a small amount of misalignment. Nevertheless, the geometric alignment can successfully resolve the remaining misalignment.

\input{Supplement/Figures/Geometric_alignment}

\section{Experimental Details}\label{appendix:details}

In this section, we present the implementation settings and detailed network architectures that are omitted from the main paper due to the space limit.
\subsection{Implementation Details}\label{appendix:implementation}
\noindent\textbf{Pose Estimation Network.}
The pose estimation network is trained by the Adam optimizer~\cite{Adamsolver} with
a weight decay of $\expnum{1}{5}$ and a learning rate set initially to $\expnum{5}{4}$ and decreased by a factor of 2 every six epochs.
Each mini-batch consists of 32 images from each lighting condition.
Following Cascaded Pyramid Network (CPN)~\cite{chen2018cascaded}, 
each human box image is cropped and resized to a fixed size, 256$\times$192.
Then we apply random scale ($0.7\sim1.35$) augmentation.
During training, we multiply a weight of $\expnum{1}{3}$ to $\mathcal{L}_\textrm{LUPI}$.

\noindent\textbf{Person Detection Network.}
The person detection network,~\ie~Cascade R-CNN~\cite{cai2018cascade} with ResNeXt101~\cite{xie2017aggregated} pre-trained on ImageNet~\cite{Imagenet}, is optimized by the SGD with a momentum of 0.9, and a weight decay of $\expnum{1}{4}$ within 12 epochs.
The initial learning rate is set to $\expnum{2}{3}$ for the detection network, and it is decayed by a factor of 0.1 for the eight and eleven epochs.  
For each lighting condition, the mini-batch size is set to 2.
All input images are resized to a fixed size, 1333$\times$800.
Similar to the pose estimation network, we multiply a weight of $\expnum{1}{3}$ to $\mathcal{L}_\textrm{LUPI}$ during training.
In testing, duplicated detected boxes are filtered out by Non-Maximum Suppression (NMS) with an IoU threshold of 0.7.

\subsection{Network Architectures}\label{appendix:networks}
Fig.~\ref{fig:pose_architecture} and Fig.~\ref{fig:detection_architecture} depict the detailed architecture of human pose estimation and person detection, respectively.

\input{Supplement/Figures/Detail_pose}
\input{Supplement/Figures/Detail_detection}

\noindent\textbf{Pose Estimation Network.}
For the human pose estimation network, as shown in Fig.~\ref{fig:pose_architecture}, we adopt CPN~\cite{chen2018cascaded}, involving two sub-networks of GlobalNet and RefineNet.
GlobalNet is similar to the feature pyramid structure for key point estimation.
In GlobalNet, each feature from four residual blocks (\ie, R1-R4) is applied 1$\times$1 convolutional kernel and then element-wise summed.
Then, RefineNet concatenates all the features from GlobalNet.
For GlobalNet and RefineNet, the pose estimation losses are applied to the output feature of each sub-network, where we denote $\mathcal{L}_{\textrm{global}}$ and $\mathcal{L}_{\textrm{refine}}$ for them, respectively.
As mentioned Sec.~5.1 in the main paper, 
LUPI is applied to the feature maps of the first convolutional layer and the following four residual blocks of a ResNet backbone.
For more details on the pose estimation network, please refer to the CPN~\cite{chen2018cascaded} paper.

\noindent\textbf{Person Detection Network.}
We utilize Cascade R-CNN~\cite{cai2018cascade} as the person detection network.
In Fig.~\ref{fig:detection_architecture}, teacher and student networks are trained by the common detection losses for classification and regression, denoted as $\mathcal{L}_{\textrm{cls}}$ and $\mathcal{L}_{\textrm{reg}}$, respectively.
We first try applying LUPI on the feature maps of the first convolutional layer and the four residual blocks following our experimental setting in human pose estimation.
However, we found that applying LUPI on the output features from the feature pyramid network is more effective than applying the loss on the features from residual blocks.
Therefore, as presented in the figure, we finally apply LUPI on the feature maps of the first convolutional layer and the following feature pyramid network.
More detailed information for the person detection network is described in the Cascade R-CNN~\cite{cai2018cascade} paper.

\section{Empirical Analysis}\label{appendix:empirical_analysis}

\subsection{Analysis on Lighting Conditions}\label{appendix:lighting_conditions}
Our method is based on the assumption that the neural style encodes the lighting condition.
In this section, we empirically verify this assumption by visualizing the distributions of Gram matrices computed from low-light and well-lit images.
Their Gram matrices are computed at 1st, 2nd, 3rd, and 4th Res Blocks of ResNet-50~\cite{resnet} pre-trained on the ImageNet dataset~\cite{Imagenet}.
\Fig{tsne} shows $t$-SNE~\cite{MaatenNov2008} visualization of Gram matrices.
In the figure, we can observe that images of the same lighting condition are grouped together in the style spaces, which validates our assumption.

\input{Supplement/Figures/Result_tsne}

\subsection{Qualitative Analysis on Lighting Condition Insensitive Features}\label{appendix:qualitative_analysis}
We conduct an experiment to investigate the impact of lighting condition insensitive features learned by our method.
To this end, we train an image reconstruction model composed of a backbone of CPN as an encoder and a decoder which consists of transposed convolution layers on the well-lit image of the ExLPose dataset.
Then, we replace the encoder of the reconstruction model with that of the pose estimation model trained by our method.
Fig.~\ref{fig:reconstruction} shows the reconstructed results on well-lit and low-light images.
The figure demonstrates that our model learns features insensitive to lighting conditions in that reconstructed images from well-lit and low-light are consistent results.

\input{Supplement/Figures/Recon_result}

\subsection{Additional Quantitative Analysis on Our Method}\label{appendix:quantitative_analysis}

In the main paper, we compare the average Hausdorff distance~\cite{huttenlocher1993comparing} between sets of Gram matrices under different lighting conditions.
In this section, we show additional results to investigate the style gaps between well-lit and paired low-light images.
To this end, we compute the mean squared error (MSE)~\cite{wang2009mean} distance on gram matrices between low-light and well-lit images of a pair before and after applying LUPI. 
Then, we report the average values for the overall ExLPose dataset.
\Fig{feature_distance} presents that the style gaps between low-light and well-lit conditions are reduced by LUPI.

\input{Supplement/Figures/MSE_distance}

\section{Results of Various Applying Enhancement Methods}\label{appendix:Variant_enhancemnt}

\subsection{Results on the ExLPose Dataset}\label{appendix:enhance_exlpose}

In this section, we report the detailed results of adopting existing enhancement methods. 
To this end, we first apply enhancement methods on low-light images of the ExLPose dataset as pre-processing.
For the learning-based enhancement method (\ie, LLFlow), we train the method using paired low-light and well-lit images of the ExLPose dataset.
For the traditional enhancement method (\ie, LIME), we apply the method on the low-light images of the ExLPose dataset without the training process.
The straightforward way to incorporate these enhancement modules with a pose estimation network is to exploit the enhanced low-light images as inputs to evaluate Baseline-well, which is the pose estimation network trained using well-lit images only.
In \Tbl{variant_enhancement}, LLFlow + Baseline-well and LIME + Baseline-well show the performance of pose estimation using LLFlow~\cite{wang2021low} and LIME~\cite{LIME_2017_TIP} as pre-processing, respectively.
They achieve inferior performance since the enhancement module is optimized to enhance the quality of low-light images but not to improve performance on the downstream recognition models.
To address this issue, we train the pose estimation network using enhanced low-light images, denoted by LLFlow + Baseline-low$^\dagger$ and LIME + Baseline-low$^\dagger$. 
The table shows that the pose estimation network trained using enhanced low-light images performs better.
We also found that training both enhanced low-light and well-lit images significantly improves the performance.
As presented in the table, LLFlow + Baseline-all$^\dagger$ and LIME + Baseline-all$^\dagger$ show the pose estimation network trained using both enhanced low-light and well-lit images achieves the best performance among other variants.
Our method outperforms all variants of adopting enhancement methods regardless of their training strategy.

\input{Supplement/Tables/Enhancement_detail}

\subsection{Results on the ExLPose-OCN Dataset}\label{appendix:enhance_exlposeocn}
We also evaluate each combination of enhancement and pose estimation methods in Sec.~\ref{appendix:enhance_exlpose} on the ExLPose-OCN dataset.
\Tbl{variant_enhancement_pidoc} shows that the tendency of each model is similar to that of each model in \Tbl{variant_enhancement}.
In detail, a trained pose estimation model using both enhanced low-light and well-lit images (\ie, LLFlow + Baseline-all$^\dagger$ and LIME + Baseline-all$^\dagger$) outperforms the variants of them.
Our method is still superior to all variants of the combinations of enhancement and pose estimation methods.

\input{Supplement/Tables/Enhancement_ocn}

\section{Additional Ablation Studies}\label{appendix:ablation}

\subsection{Effect of LSBN}\label{appendix:effect_LSBN}

This section presents extensive experiments to investigate the effect of LSBN.
As mentioned Sec.~5.1.1 in the main paper, domain adaptation (DA) methods cannot effectively reduce the large domain gap between low-light and well-lit conditions.
As presented in Table~\ref{tab:ablation_da}, DANN~\cite{dann}, AdvEnt~\cite{vu2019advent}, and LUPI show inferior performance, proving the necessity of LSBN.
When they are combined with LSBN, the performance of each method is improved since it successfully bridges the large domain discrepancy between low-light and well-lit conditions.
When compared with `LSBN + DANN' and `LSBN + AdvEnt', our method (\ie, `LSBN + LUPI') outperforms them thanks to the effectiveness of our neural style-based approach, LUPI.
It is worth noting that our method can be a plug-and-play to the feature extractor, while AdvEnt is hard to apply to the task where the entropy map cannot be computed.

\input{Supplement/Tables/LSBN_effect}

\subsection{Effect of the Neural Style of LUPI}\label{appendix:feature_regression}
We investigate the effect of LUPI by comparing LSBN + LUPI (\ie, our method) with LSBN + LUPI-\textit{feat} that directly approximates feature maps of the teacher instead neural styles.
We conduct an in-depth analysis by comparing them in terms of the feature gaps between low-light and well-lit conditions.
As presented in Fig.~\ref{fig:lupi_effect}, LSBN + LUPI effectively reduces the average Hausdorff distance of feature maps between different lighting conditions, although LSBN + LUPI-\textit{feat} directly approximates the feature maps of well-lit images.
We conjecture that the advantage of using neural styles over directly using features comes from the less image-dependent characteristic of neural styles, \ie, features are more image-specific information that changes with each image; thus it is more difficult for LSBN + LUPI-\textit{feat} to reduce the difference between features.
In consequence, LSBN + LUPI effectively aligns the feature distributions of different lighting conditions since they aim to approximate the styles which represent the characteristics of lighting conditions.

\input{Supplement/Figures/Feature_distance}

\subsection{Effect of Layer Selection of LUPI}\label{appendix:layer_selection}

We investigate the impact of the selection of layers where LUPI is applied.
Table~\ref{tab:layerablation} compares different choices for the layer selection.
In the table, C1 applies LUPI to the output of the first convolutional layer,
and C1:R$n$ applies LUPI to the 1st to the $n$-th residual blocks as well as the first convolution layer.
As shown in the table, the performance improves as LUPI is applied to more blocks,
and our final model (C1:R4) achieves the best performance.

\input{Supplement/Tables/Layer_ablation}

\subsection{Effect of the Gradient Direction of LUPI}\label{appendix:gradient}
When training with LUPI, we let the gradient from the loss flow \emph{only} to the student in order to train the student with privileged information from the teacher, i.e., information flows in one direction from the teacher to the student.
In this section, we study the effect of this one-direction strategy on LUPI.
To this end, we prepare three variants of our approach: `T $\rightarrow$ S (Ours)', `T $\leftrightarrow$ S' and `T $\leftarrow$ S'.
`T $\rightarrow$ S (Ours)' is our proposed approach.
`T $\leftrightarrow$ S' allows the gradient from LUPI to flow to both the teacher and student models, i.e., the teacher and student can affect each other.
`T $\leftarrow$ S', on the other hand, allows the gradient to flow only to the teacher.
Table~\ref{tab:lupi_loss} compares the performance of these three variants.
As shown in the table, our approach clearly outperforms the others.
This result implies our one-direction strategy is essential for learning about LUPI.

\input{Supplement/Tables/Lupi_direction}

\subsection{Effect of Intensity Scaling}\label{appendix:scaling}

As described in the main paper, the average channel intensity of each low-light image is automatically scaled to 0.4 before being fed to the student network.
\Tbl{scaling_effect} shows the performance of Baseline-all and the proposed method trained on original low-light images and scaled low-light images.
In low-light conditions, automatically scaled low-light images significantly improve the performance of both models. However, Baseline-all trained on the scaled low-light images performs much worse in well-lit conditions.

We suspect that, in the case of using original low-light images, the Baseline-all model is biased to the well-lit condition.
It is because well-lit images have large pixel intensities, so the scale of gradient of them is larger than that of original low-light images.
Then, in the case of using scaled low-light images, the Baseline-all model is less biased for the well-lit condition, so the performance on the well-lit condition is decreased.
However, the proposed method is less biased due to the lighting condition invariant features of LSBN and LUPI.
Consequently, Table~\ref{tab:scaling_effect} demonstrates that intensity scaling of low-light images improves the performance of both Baseline-all and our method for low-light conditions.

\input{Supplement/Tables/Scaling_effect}

\section{Results of Person Detection}\label{appendix:detection}
The ExLPose dataset provides human pose and bounding box labels for training and evaluation of human pose estimation methods.
Moreover, human bounding boxes in the ExLPose dataset can serve as a detection dataset on low-light images. 
We adopt Cascade R-CNN~\cite{cai2018cascade} as our person detection network and compare our method with other solutions in the same way as described in the main paper.

\Tbl{Human_Detection} shows summarized the person detection performance of ours and other solutions.
In the table, Baseline-all is a Cascade R-CNN trained on both low-light and well-lit images with person detection loss only, and Baseline-all still underperforms our method.
For comparing other solutions, we adopt LLFlow and LIME for enhancement methods and DANN for domain adaptation. 
AdvEnt cannot be applied to the person detection task as the method is based on the entropy minimization of prediction. Accordingly, we did not conduct experiments about AdvEnt on person detection.
As shown in the table, these methods rather degrade performance in low-light conditions as a direct adoption of enhancement and domain adaptation is not effective.
On the other hand, our method outperforms by large margins in all the low-light conditions, and the results show the effectiveness of our method for person detection in the low-light condition.

\input{Supplement/Tables/Human_detection}

\section{Additional Qualitative Results}\label{appendix:additional_qual}

\subsection{Results on the ExLPose-OCN Dataset}\label{appendix:qualatitive_ExLPose-OCN}

\input{Supplement/Figures/ExLPoseOCN}

We provide the ExLPose-OCN dataset to evaluate the generalization capability of pose estimation methods to unseen cameras.
\Fig{qualitative_results_ExLPose_OCN} shows qualitative comparisons of our method with Baseline-all, DANN, and LIME + Baseline-all on the ExLPose-OCN dataset.
As shown in the figure, our method accurately estimates human poses, but other methods do not generalize well to unseen cameras and often fail to estimate accurate human poses.

\subsection{Results on the ExLPose Dataset}

\Fig{qual} shows additional qualitative results of Baseline-all, DANN~\cite{dann}, LIME~\cite{LIME_2017_TIP} + Baseline-all and our method. 
This again demonstrates that Baseline-all and DANN often fail to predict poses, while our method surpasses them.

\subsection{Failure Cases of Our Method}
We provide failure cases of our method in \Fig{failure}. 
The first row of the figure shows the results of the pose estimation network on low-light images which have little pixel information.
In such images, noise components are prevalent, and the remaining pixel information is too small to estimate human poses.
Our method also fails to predict human poses for occluded humans, as shown second and third rows in the figure. 

\subsection{Results of Enhancement Methods}

\Fig{enhance} shows enhanced low-light images of the ExLPose and ExLPose-OCN datasets using LLFlow~\cite{wang2021low} and LIME~\cite{LIME_2017_TIP}.
For the ExLPose dataset, LLFlow successfully enhances low-light images and reduces the noises of low-light images.
However, LLFlow cannot generalize well to the ExLPose-OCN dataset due to different image signal processors and exhibit different noise distributions.
Enhanced low-light images using LIME have the remaining noise as the method does not consider noise well.
These limitations may reduce the generalization capability and performance of the pose estimation when enhancement methods are combined with the pose estimation network.

\subsection{Results of Multi-person Pose Estimation}
\Fig{detection_qual} presents qualitative results for the person detection of Baseline-all, DANN, LIME + Baseline-all, and our method.
These predicted bounding boxes of our method are exploited for multi-person pose estimation; its qualitative results are shown in \Fig{multipose_qual}. 
The figure shows that our method successfully performs multi-person pose estimation while other solutions largely fail to estimate human poses. 
\clearpage

\input{Supplement/Figures/Other_qual}

%% file: Supplement/Figures/SID_ExLPose.tex
\begin{figure}[ht]
    \centering
    \includegraphics[width=1.0\linewidth]{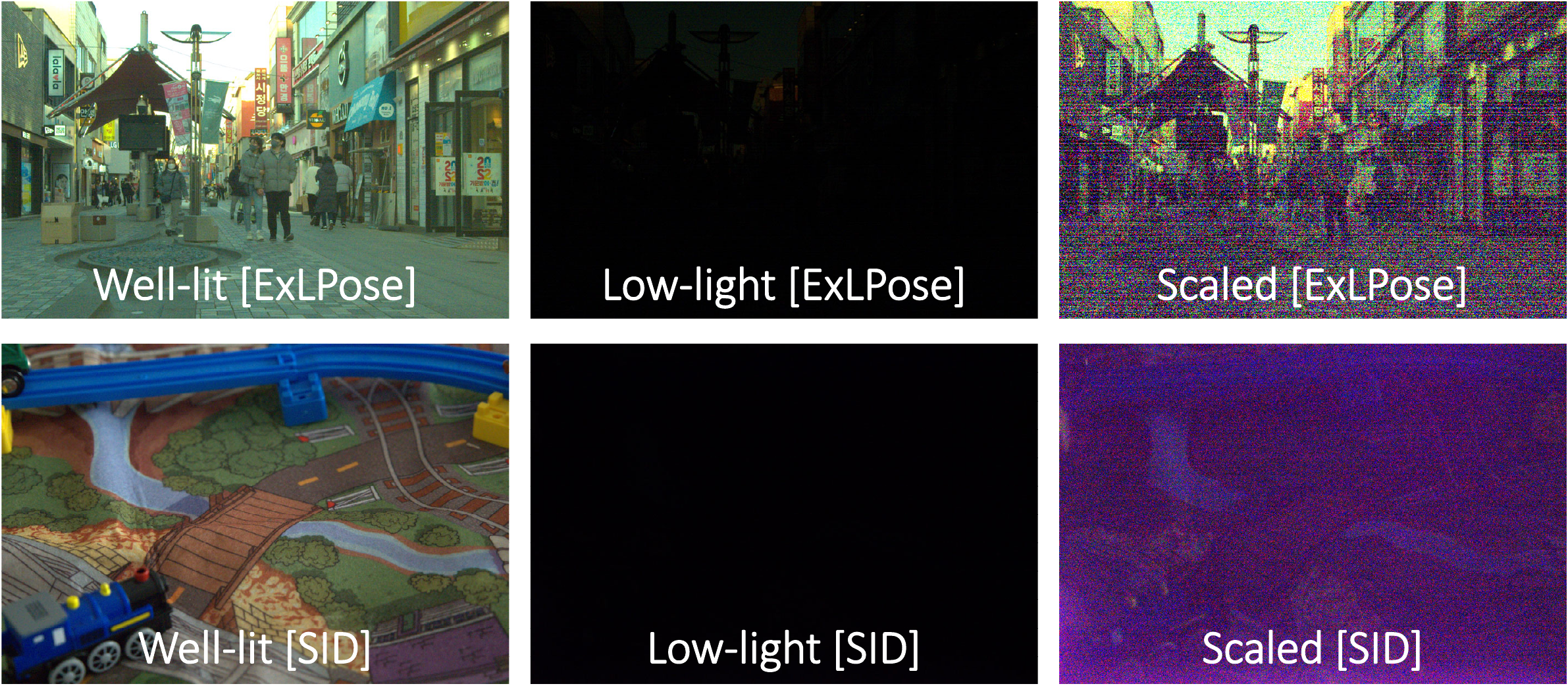}
    \vspace{-3mm}
\caption{
Examples of well-lit, extremely low-light, and scaled low-light images in the SID~\cite{chen2018learning} and ExLPose datasets. 
} \label{fig:SID_and_ExLpose}
\end{figure}

%% file: Supplement/Tables/Dataset_scale.tex
\begin{table}[h]
\centering
\vspace{0mm}
\renewcommand{\arraystretch}{1.0}
\scalebox{0.96}{\begin{tabular*}{1.04\columnwidth}{lcccc}
\toprule
\centering
Dataset  & Paired & \#Poses  & Multi-person  \\ \midrule
      LSP~\cite{johnson2010clustered} & & 2,000  &   \\
      LSP Extended~\cite{johnson2011learning} & & 10,000  &   \\
      MPII Single-person~\cite{andriluka20142d} & & 26,429  &    \\
      FLIC~\cite{sapp2013modec} & & 5,003  &    \\\midrule
      We are family~\cite{eichner2010we}  &  &  3,131  &  \checkmark      \\
      OCHuman~\cite{zhang2019pose2seg} &  & 8,110 & \checkmark    \\
      MPII Multi-person~\cite{andriluka20142d} &  & 14,993 & \checkmark    \\
      CrowdPose~\cite{li2019crowdpose} &  & 80,000 & \checkmark    \\
      \rowcolor[gray]{.8} ExLPose (Ours)    & \checkmark  & 14,214 &  \checkmark  \\
\bottomrule
\end{tabular*}}
\vspace{-2mm}
\caption{Statistics of the ExLPose and other datasets.}

\vspace{-1mm}
\label{tab:statistics_datasets}
\end{table}

%% file: Supplement/Figures/Geometric_alignment.tex
\begin{figure}[h]
    \centering
    \scalebox{1.0}{
    \includegraphics[width=\linewidth]{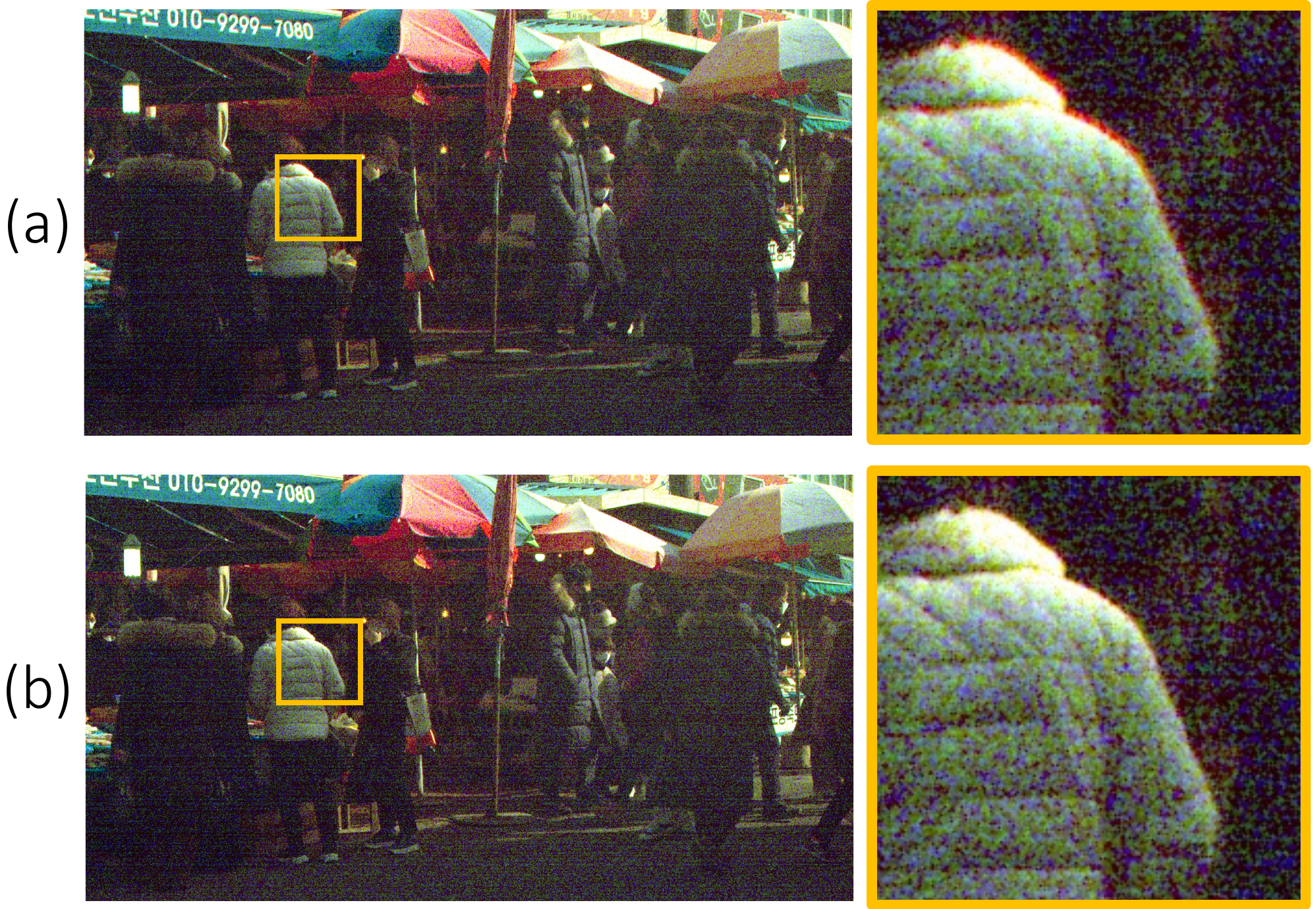} }
    \vspace{-4mm}
\caption{Stereo-anaglyph images (a) before and (b) after geometric alignment.
} \label{fig:geometric_alignment}
\end{figure}

%% file: Supplement/Figures/Detail_pose.tex
\begin{figure*}[t]
    \centering
    \includegraphics[width=0.92\linewidth]{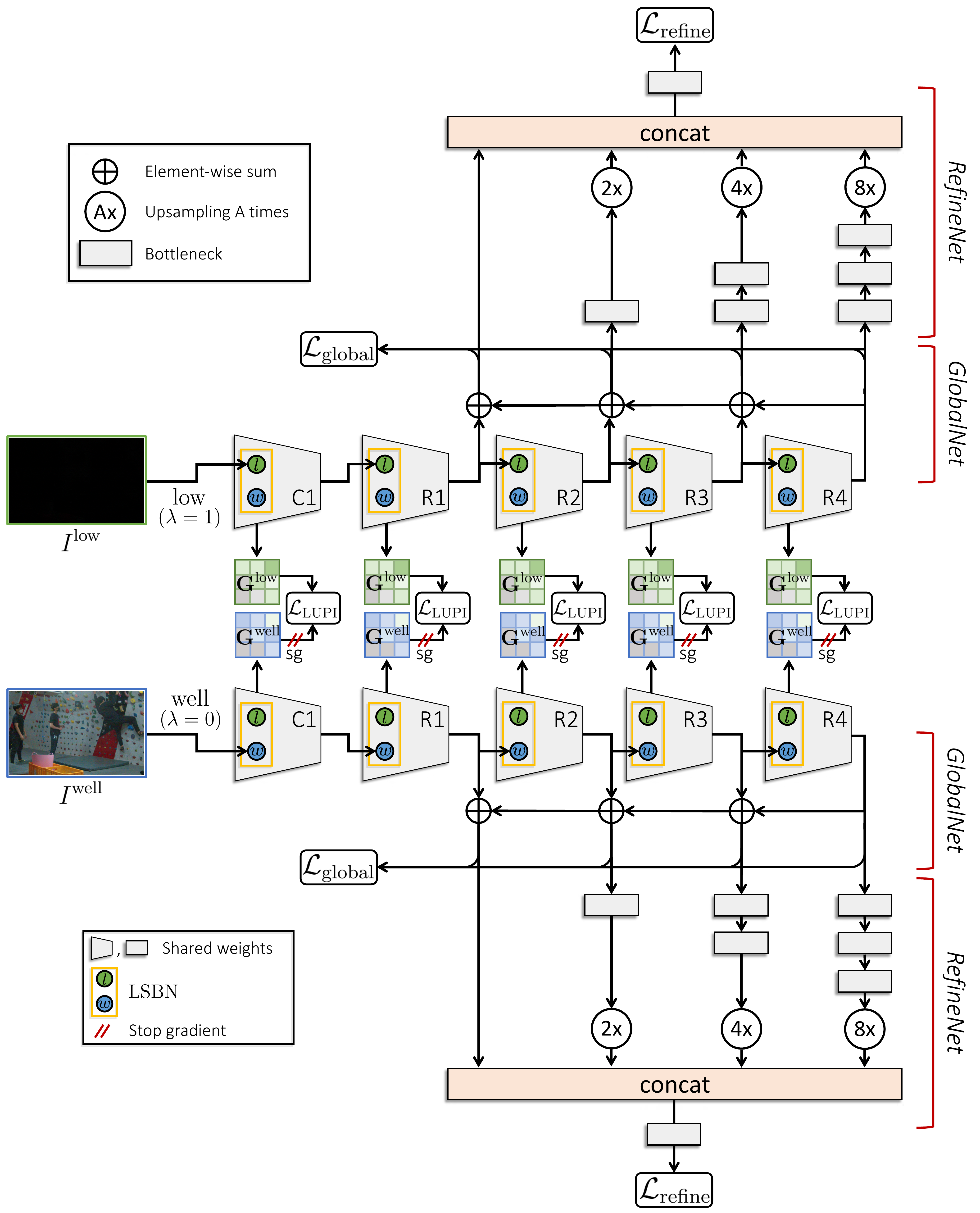}
\caption{Illustration of the detailed human pose estimation network architecture. Both of teacher (\textit{bottom}) and student (\textit{top}) are trained by the same pose estimation loss of $\mathcal{L}_{\textrm{global}}$ and $\mathcal{L}_{\textrm{refine}}$ from GlobalNet and RefineNet of CPN~\cite{chen2018cascaded}.
LUPI is applied to the feature maps of the first convolutional layer (\ie, C1) and four residual blocks (\ie, R1-R4) of a ResNet backbone.
Teacher and student share all the parameters except LSBNs. 
} \label{fig:pose_architecture}
\end{figure*}

%% file: Supplement/Figures/Detail_detection.tex
\begin{figure*}[t]
    \centering
    \vspace{-1mm}
    \includegraphics[width=\linewidth]{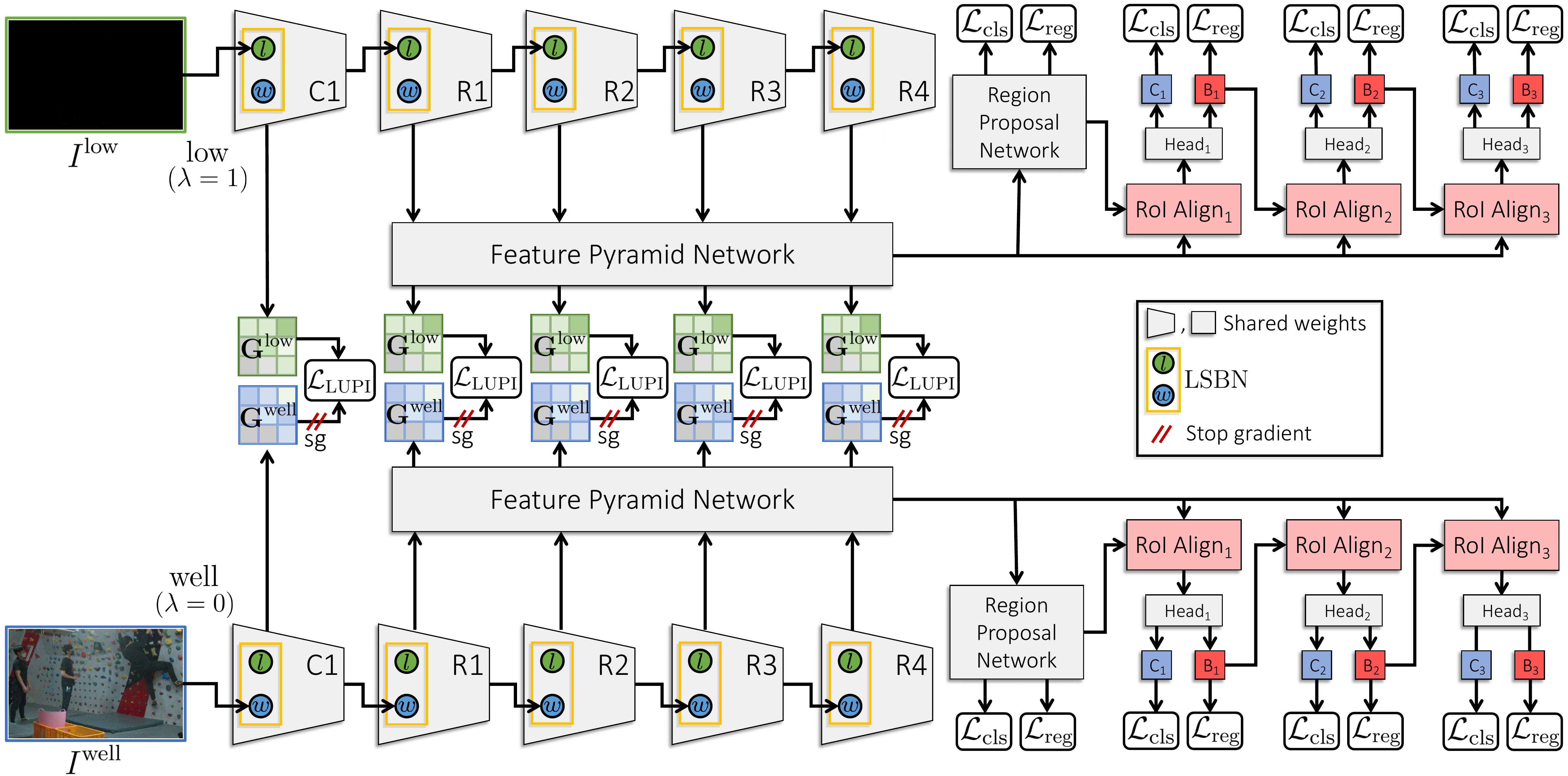}
    \vspace{-4mm}
\caption{The overall architecture of the person detection network and training strategy.
Both of teacher (\textit{bottom}) and student (\textit{top}) are trained by the same detection loss functions ($\mathcal{L}_{\textrm{cls}}$ and $\mathcal{L}_{\textrm{reg}}$), and student takes additional supervision from teacher through LUPI. 
The loss for LUPI is applied to the feature maps of the first convolutional layer (\ie, C1) and the following feature pyramid network (FPN) which takes feature maps of four residual blocks (\ie, R1-R4) of a ResNet backbone as inputs. 
Teacher and student share all the parameters except LSBNs. 
``Head'', ``C'', and ``B'' denote roi head, object classification, and bounding box, respectively.
} \label{fig:detection_architecture}
\end{figure*}

%% file: Supplement/Figures/Result_tsne.tex
\begin{figure*}[t]
    \centering
    \vspace{0mm}
    \scalebox{1}{
    \includegraphics[width=\linewidth]{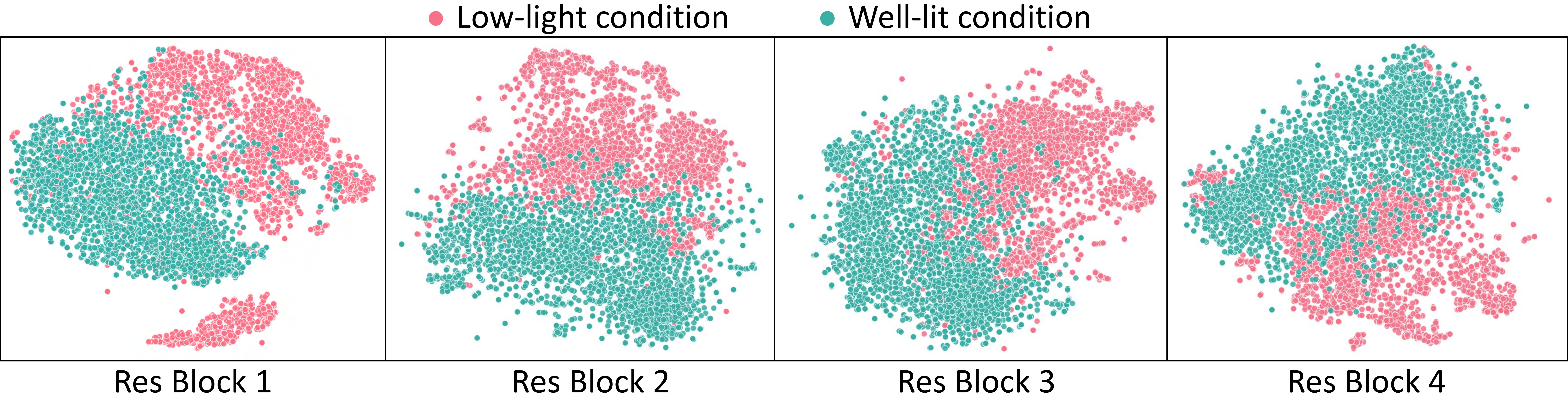} }
\caption{
$t$-SNE visualization of the distributions of Gram matrices computed from low-light and well-lit images.
The Gram matrices are computed from feature maps of the 1st, 2nd, 3rd, and 4th Res Blocks of ResNet-50~\cite{resnet} pre-trained on the ImageNet dataset~\cite{Imagenet}.
In all the visualizations, images of the same lighting condition are clustered together, suggesting that neural styles in the form of Gram matrices encode lighting conditions.
} \label{fig:tsne}
\vspace{0mm}
\end{figure*}

%% file: Supplement/Figures/Recon_result.tex
\begin{figure*}[t]
    \centering
    \vspace{0mm}
    \scalebox{1}{
    \includegraphics[width=\linewidth]{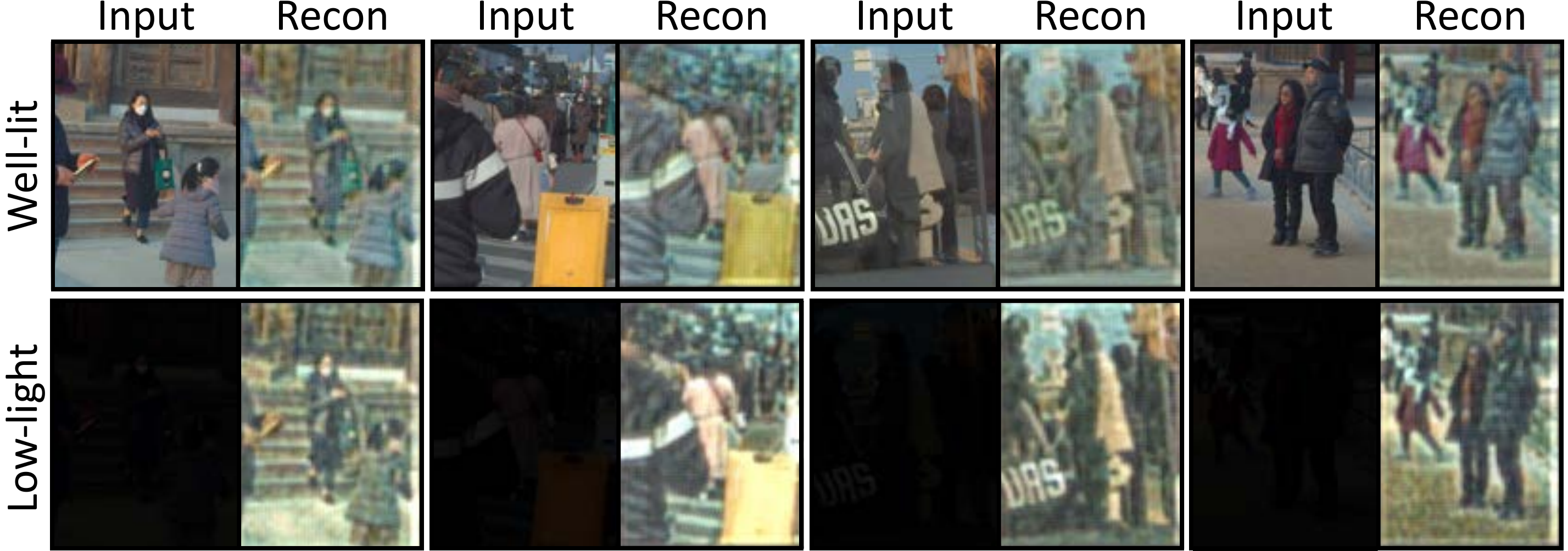} }
\caption{
Input images and reconstructed images by the proposed method.
} \label{fig:reconstruction}
\vspace{0mm}
\end{figure*}

%% file: Supplement/Figures/MSE_distance.tex
\begin{figure}[h]
    \centering
    \vspace{0mm}
    \scalebox{0.85}{
    \includegraphics[width=\linewidth]{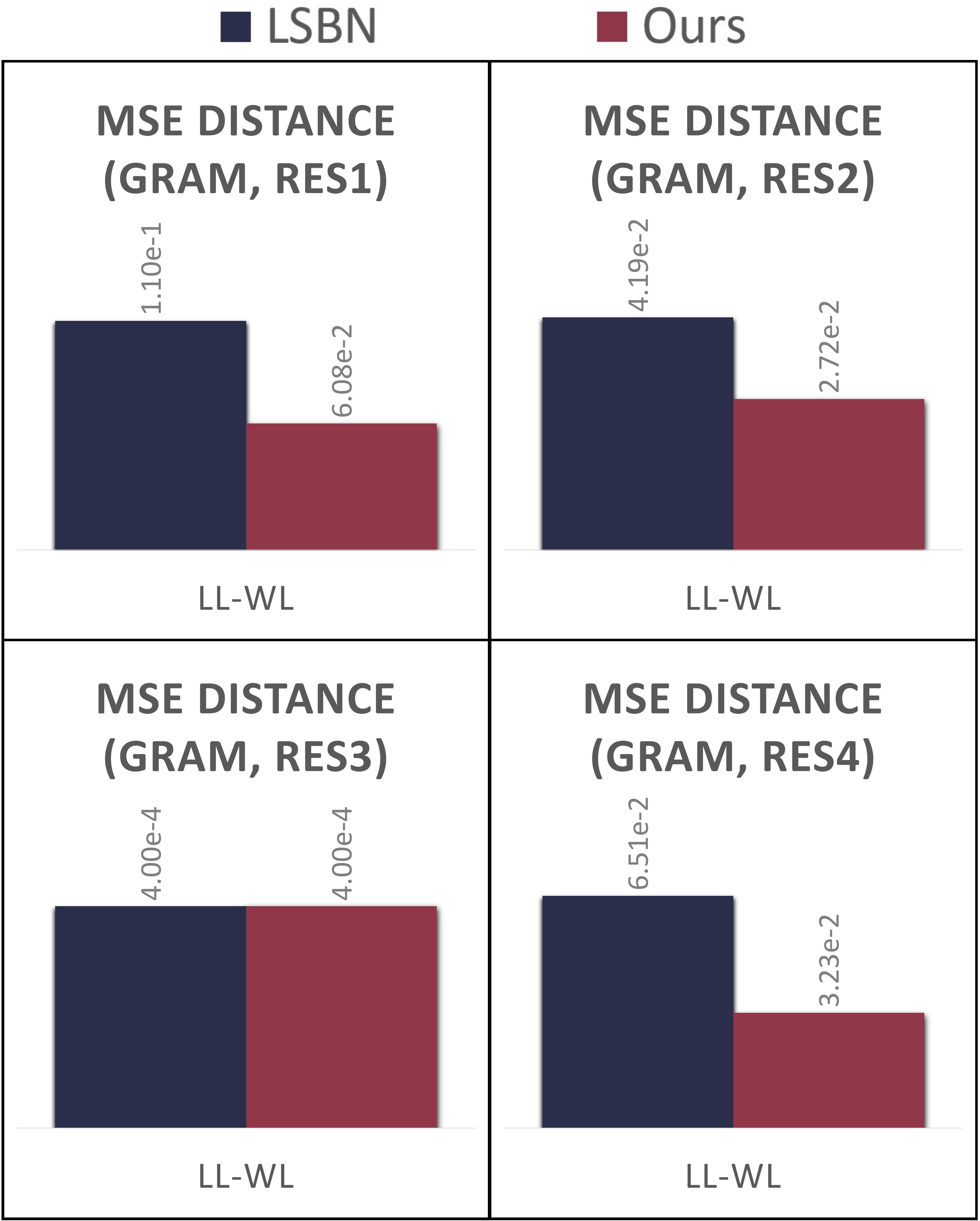} }
\caption{
Style gaps between low-light (LL) and well-lit (WL) conditions.
The gap is computed by averaging the mean squared error (MSE) distance between low-light and paired well-lit images on the ExLPose dataset.
} \label{fig:feature_distance}
\end{figure}

%% file: Supplement/Tables/Enhancement_detail.tex
\begin{table}[h]
\centering
\renewcommand{\arraystretch}{1.0}
\scalebox{0.85}{\begin{tabular}{lccccc}
\toprule
AP@0.5:0.95     & LL-N     &  LL-H       &  LL-E   & LL-A  &  WL      \\\midrule
Baseline-well    &    23.5        &    7.5       &    1.1     &  11.5    & \textbf{68.8} \\
Baseline-low      &    32.6      &     25.1      &    13.8     &  24.6    & 1.6  \\
Baseline-all     &   33.8        &   	25.4        & 	\underline{14.3}	 &  25.4       &   57.9  \\\midrule
LLFlow + Baseline-well     &  22.9     &    11.7	    &   2.4   & 12.8  & -   \\
LLFlow + Baseline-low$^\dagger$      & 30.7	   &   18.4		    &   8.0	   & 19.6 &	40.9  \\
LLFlow + Baseline-all$^\dagger$ & 35.2   &    	20.1	    &	8.3	   & 22.1  & 	65.1  \\\midrule
LIME + Baseline-well      & 23.1 &   5.8  &   1.0	  & 10.8  & -  \\
LIME + Baseline-low$^\dagger$                 & 31.6 & 24.3	& 12.6 & 23.6 & 36.2  \\
LIME + Baseline-all$^\dagger$                 & \underline{40.6}	& \underline{27.1} &	13.4 & \underline{28.3}	  & 63.2 \\ \midrule
Ours & \textbf{42.3} & \textbf{34.0} & \textbf{18.5} & \textbf{32.7} & \underline{68.5}  \\ \bottomrule
\end{tabular}}
\caption{
Quantitative results in AP@0.5:0.95 on the ExLPose dataset;
Low-light-normal, Low-light-hard, Low-light-extreme, Low-light-all, Well-lit splits.
Baseline-low$^\dagger$ is a model trained using enhanced low-light images, and Baseline-all$^\dagger$ denotes a model trained using both enhanced low-light and well-lit images.}
\label{tab:variant_enhancement}
\end{table}

%% file: Supplement/Tables/Enhancement_ocn.tex
\begin{table}[h]
\centering
\renewcommand{\arraystretch}{1.0}
\scalebox{0.93}{\begin{tabular}{lccc}
\toprule
AP@0.5:0.95       &  A7M3  &  RICOH3 &  Avg.    \\\midrule
Baseline-well  & 23.7    & 23.9 &  23.8 \\
Baseline-low     & 15.2  & 15.6 &  15.4\\
Baseline-all    &  32.8 & \underline{31.7} &  \underline{32.2}   \\\midrule
LLFlow + Baseline-well  & 30.4 & 24.5 &  27.3  \\
LLFlow + Baseline-low$^\dagger$   & 20.5 &  18.7 &  19.5 \\
LLFlow + Baseline-all$^\dagger$   &  25.6  &  28.2 &  27.0 \\\midrule
LIME + Baseline-well    & 10.9 &  7.9 &  9.3\\
LIME + Baseline-low$^\dagger$     & 20.7 & 13.4 &  16.8 \\
LIME + Baseline-all$^\dagger$     & \underline{33.2} & 28.4 &  30.7 \\ \midrule
Ours    & \textbf{35.3}   &  \textbf{35.1} &  \textbf{35.2} \\ \bottomrule
\end{tabular}}
\caption{
Quantitative results in AP@0.5:0.95 on the ExLPose-OC dataset;
A7M3, and RICOH3 splits.
Baseline-low$^\dagger$ is a model trained using enhanced low-light images, and Baseline-all$^\dagger$ denotes a model trained using both enhanced low-light and well-lit images.}
\label{tab:variant_enhancement_pidoc}
\end{table}

%% file: Supplement/Tables/LSBN_effect.tex
\begin{table}[h]
\centering
\renewcommand{\arraystretch}{1.0}
\scalebox{0.9}{\begin{tabular}{lcccccc}
\toprule
AP@0.5:0.95     & LL-N     &  LL-H       &  LL-E   & LL-A  &  WL \\\midrule
DANN & 34.9	   & 	24.9	 & 	13.3		 & 25.4   &	58.6    	       \\
AdvEnt & 	33.0  & 	24.1	 &	11.6  &	23.8   &	60.0  	  \\
LUPI & 	34.2  & 	23.1	 &	11.2  &	24.0   &	61.7  	  \\ \midrule
LSBN + DANN     &  \underline{42.2}   &  	30.5		  &  16.7  &  30.8	& \underline{67.4}     \\
LSBN + AdvEnt   & 41.3	& \underline{31.2}	   & \textbf{19.0}    & \underline{31.5} & \textbf{68.5}    \\ 
LSBN + LUPI (Ours) & \textbf{42.3}   & \textbf{34.0}   & \underline{18.5}       &  \textbf{32.7}  &	\textbf{68.5}     \\ \bottomrule
\end{tabular}}
\caption{
Analysis for the effect of LSBN. 
The results are reported in AP@0.5:0.95 on the ExLPose dataset;
Low-light-normal, Low-light-hard, Low-light-extreme, Low-light-all, and Well-lit conditions.}
\label{tab:ablation_da}
\end{table}

%% file: Supplement/Figures/Feature_distance.tex
\begin{figure}[h]
    \centering
    \vspace{0mm}
    \scalebox{0.9}{
    \includegraphics[width=\linewidth]{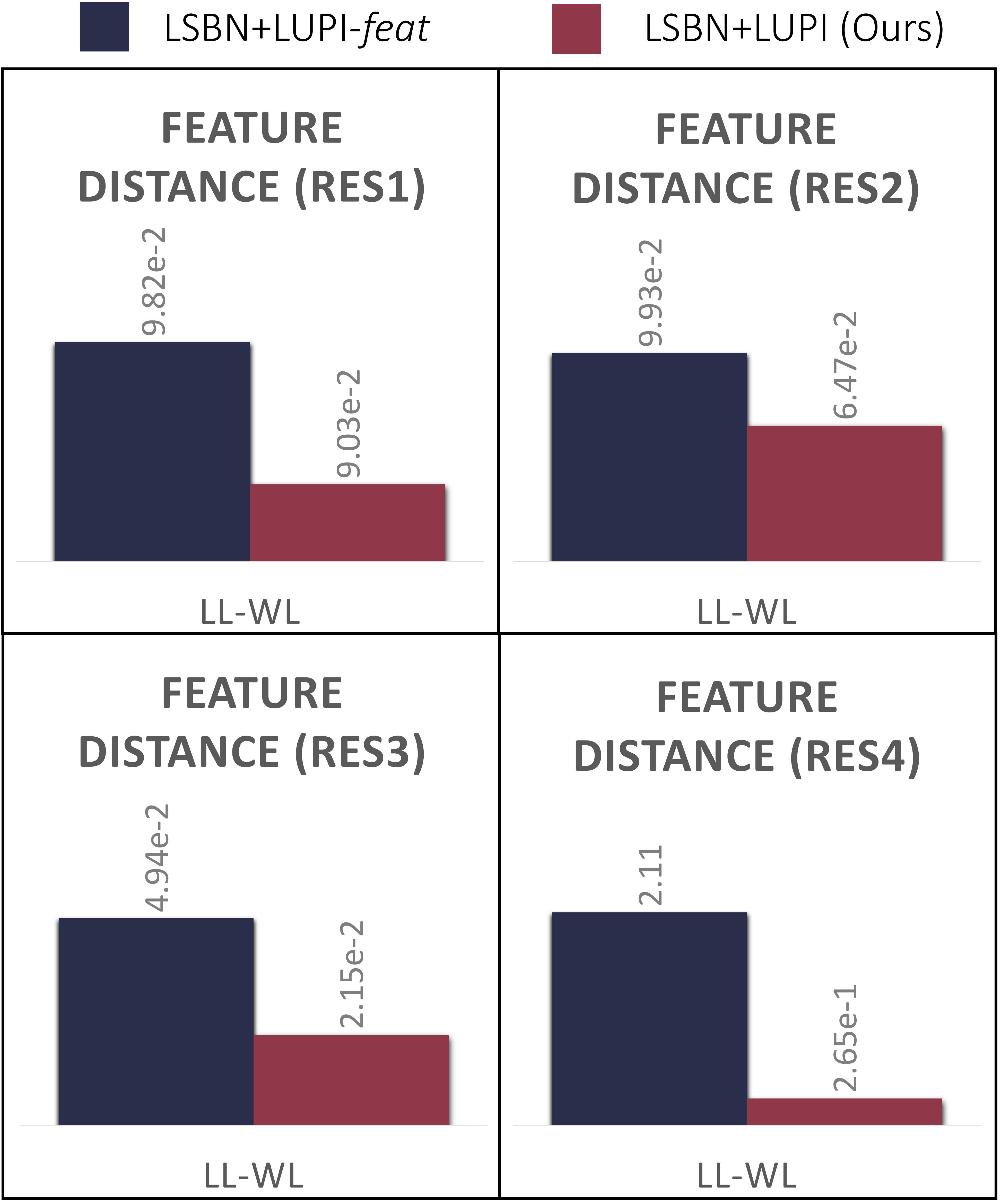} }
    \vspace{-2mm}
\caption{
Quantitative analysis on feature distribution gap between low-light (LL) and well-lit (WL) conditions.
The distance is measured by the average Hausdorff distance between lighting condition sets.
} \label{fig:lupi_effect}
\vspace{0mm}
\end{figure}

%% file: Supplement/Tables/Layer_ablation.tex
\begin{table}[h]
\centering
\renewcommand{\arraystretch}{1.0}
\scalebox{1}{
\begin{tabular}{lccccc}
\toprule
AP@0.5:0.95       & LL-N     &  LL-H       &  LL-E   & LL-A  &  WL     \\ \midrule
C1        &  41.0	   &  30.0	    &  15.0		  & 29.7	  &     67.6    \\
C1:R1    &    40.8	   &   30.0      &   17.0    &   30.3      & 	67.1   \\
C1:R2      &  42.2	  &   30.6   &	15.8    &	30.7    & \underline{67.9}         \\
C1:R3    & \textbf{43.1}	      & \underline{32.2}     & 	\underline{17.7}    &   \underline{32.2}	& \underline{67.9}           \\
Ours (C1:R4)   & \underline{42.3}	   & \textbf{34.0}	 & \textbf{18.5} &	\textbf{32.7} & \textbf{68.5} 	 	       \\ \bottomrule
\end{tabular}
}
\vspace{-2mm}
\caption{
Analysis on layers where LUPI is applied.}
\label{tab:layerablation}
\end{table}

%% file: Supplement/Tables/Lupi_direction.tex
\begin{table}[h]
\centering
\vspace{0mm}
\renewcommand{\arraystretch}{1.0}
\scalebox{1}{\begin{tabular}{lccccccc}
\toprule
AP@0.5:0.95      & LL-N     &  LL-H       &  LL-E   & LL-A  &  WL        \\ \midrule
T $\leftrightarrow$ S    & \underline{41.6}		    &  \underline{32.8}		  &  \underline{16.7}	 & \underline{31.6}	 &    67.0  \\
T $\leftarrow$ S    &  39.7	 & 	29.6	   & 15.7	  & 29.5	 &    \underline{68.4}  \\
T $\rightarrow$ S (Ours)    & \textbf{42.3}	 & \textbf{34.0}	    & \textbf{18.5}	  &  \textbf{32.7}	 &  \textbf{68.5}       \\ \bottomrule
\end{tabular}}
\vspace{-2mm}
\caption{
Analysis on the impact of the gradient direction from LUPI.}
\label{tab:lupi_loss}
\end{table}

%% file: Supplement/Tables/Scaling_effect.tex
\begin{table}[h]
\centering
\vspace{0mm}
\renewcommand{\arraystretch}{1.0}
\scalebox{0.80}{\begin{tabular}{llccccc}
\toprule
 AP@0.5:0.95  & Method   & LL-N     &  LL-H       &  LL-E   & LL-A   &  WL        \\ \midrule
\multirow{2}{*}{No scaling} & Baseline-all       &	22.3  & 6.5	& 2.5	&11.2	 & 61.3   \\
& Ours       & 25.5	  & 	10.0  &  5.9  & 14.5 &	\underline{67.1}       \\ \midrule
\multirow{2}{*}{Scaling}  & Baseline-all    &   \underline{33.8}        &   	\underline{25.4}        & 	\underline{14.3}	 &  \underline{25.4}       &   57.9          \\
 & Ours      & \textbf{42.3}   & \textbf{34.0}   & \textbf{18.5}       &  \textbf{32.7}  &	\textbf{68.5}   \\  \bottomrule
\end{tabular}}
\vspace{-2mm}
\caption{
Analysis on impact of scaling for low-light images.}
\vspace{-1mm}
\label{tab:scaling_effect}
\end{table}

%% file: Supplement/Tables/Human_detection.tex
\begin{table}[h]
\centering
\renewcommand{\arraystretch}{1.0}
\scalebox{0.88}{\begin{tabular}{lcccccc}
\toprule
AP@0.5:0.95   & LL-N     &  LL-H       &  LL-E   & LL-A  &  WL \\\midrule
Baseline-low    & 39.8  & 30.5  & 17.4  & 30.2  & 40.7 \\
Baseline-well   & 22.6  & 3.8   & 0.5   & 9.7   & 53.2 \\
Baseline-all    & 44.9  & \underline{32.3}  & \underline{18.3}  & \underline{33.0}  & 60.8 \\\midrule
LLFlow + Baseline-all   & 38.3  & 29.0  & 15.2  & 28.4  & \underline{61.3} \\
LIME + Baseline-all     & \underline{45.8}  & 32.1  & 17.0  & 32.8  & \textbf{63.5} \\ \midrule
DANN      & 43.7  & 30.1  & 14.8  & 30.7  & 55.7  \\ \midrule
Ours            & \textbf{46.2}  &\textbf{34.5}  & \textbf{21.0}  & \textbf{34.9}  & 60.9 \\ \bottomrule
\end{tabular}}
\vspace{-2mm}
\caption{
Person detection results in AP@0.5:0.95 on the ExLPose dataset;
Low-light-normal, Low-light-hard, Low-light-extreme, Low-light-all, and Well-lit conditions.}
\vspace{-2mm}
\label{tab:Human_Detection}
\end{table}

%% file: Supplement/Figures/ExLPoseOCN.tex
\begin{figure}[ht]
    \centering
    \includegraphics[width=0.95\linewidth]{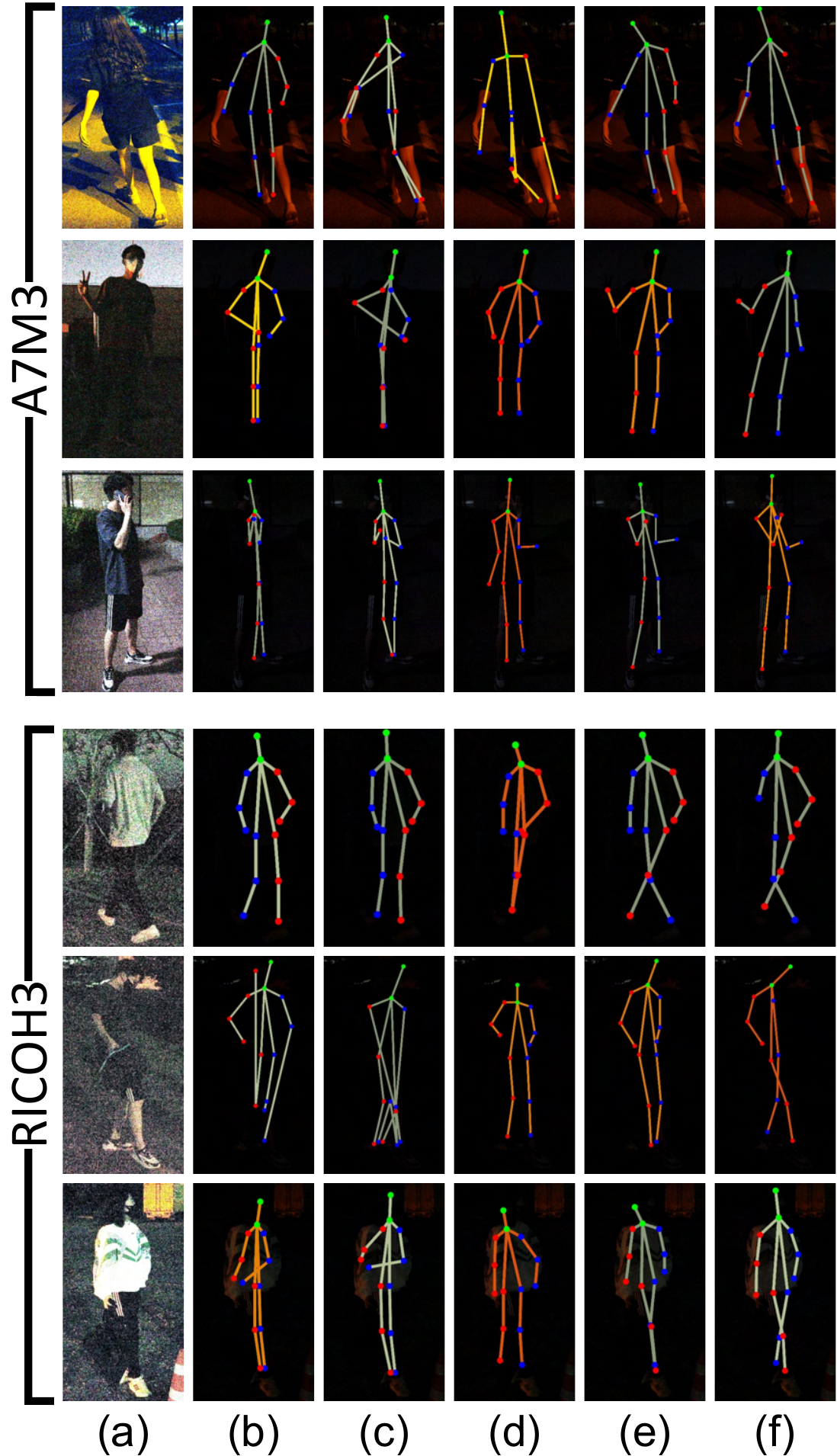}
    \vspace{-2mm}
\caption{
Qualitative results on the ExLPose-OCN dataset. 
Predicted poses and labels are visualized on corresponding low-light images.
(a) Scaled low-light images. (b) Baseline-all. (c) DANN. (d) LIME + Baseline-all. (e) Ours. (f) Ground-truth.
} \label{fig:qualitative_results_ExLPose_OCN}
\vspace{-2mm}
\end{figure}

%% file: Supplement/Figures/Other_qual.tex
\begin{figure*}[t]
    \centering
    \vspace{-2mm}
    \scalebox{1}{
    \includegraphics[width=1.0\linewidth]{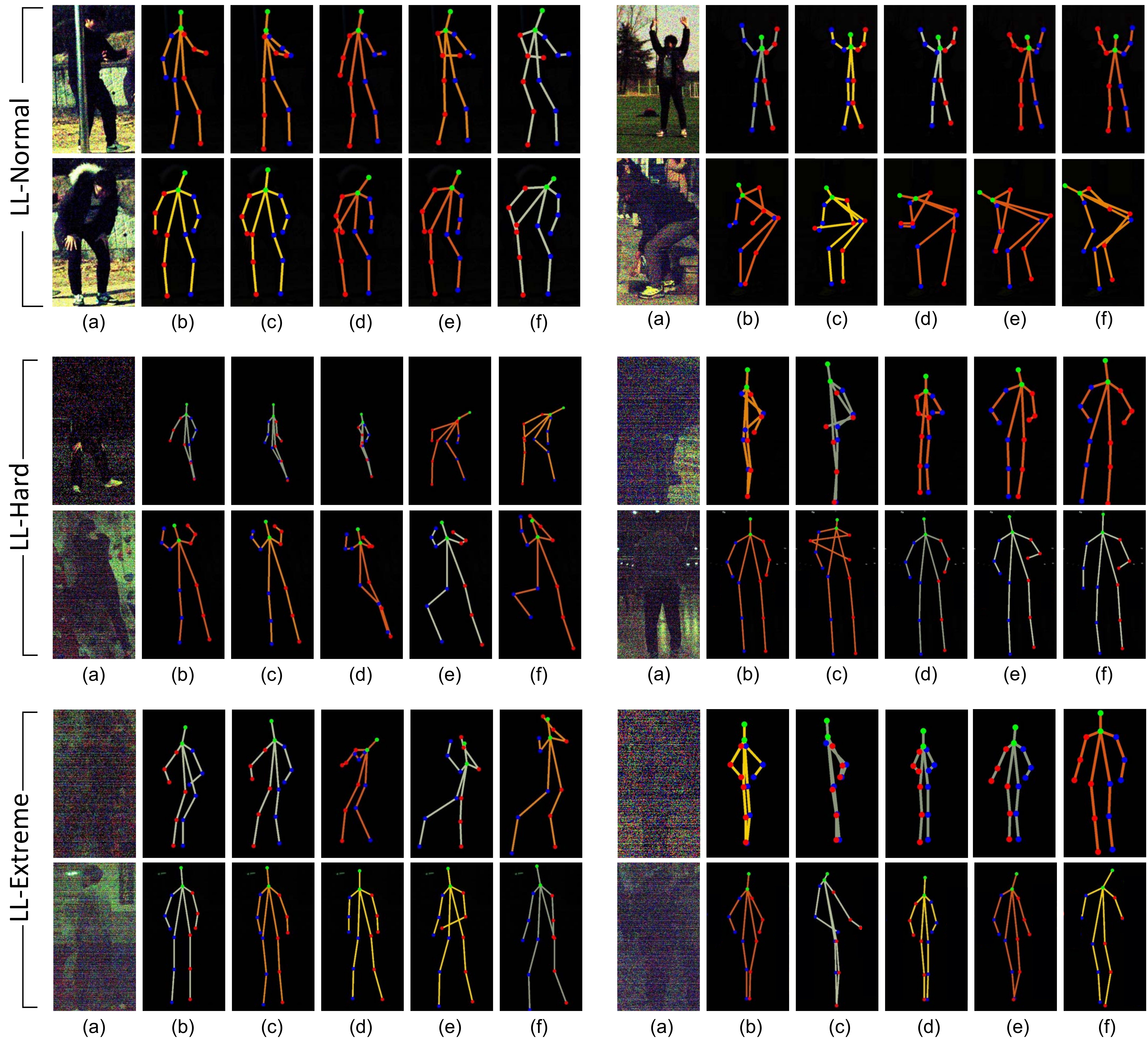} }
    \vspace{-3mm}
\caption{
Qualitative results of single-person pose estimation on the ExLPose dataset.
Predicted poses and labels are visualized on corresponding low-light images.
(a) Scaled low-light images. (b) Baseline-all. (c) DANN. (d) LIME + Baseline-all. (e) Ours. (f) Ground-truth.
} \label{fig:qual}
\vspace{-6mm}
\end{figure*}

\begin{figure*}[t]
    \centering
    \vspace{-2mm}
    \scalebox{1.0}{
    \includegraphics[width=0.85\linewidth]{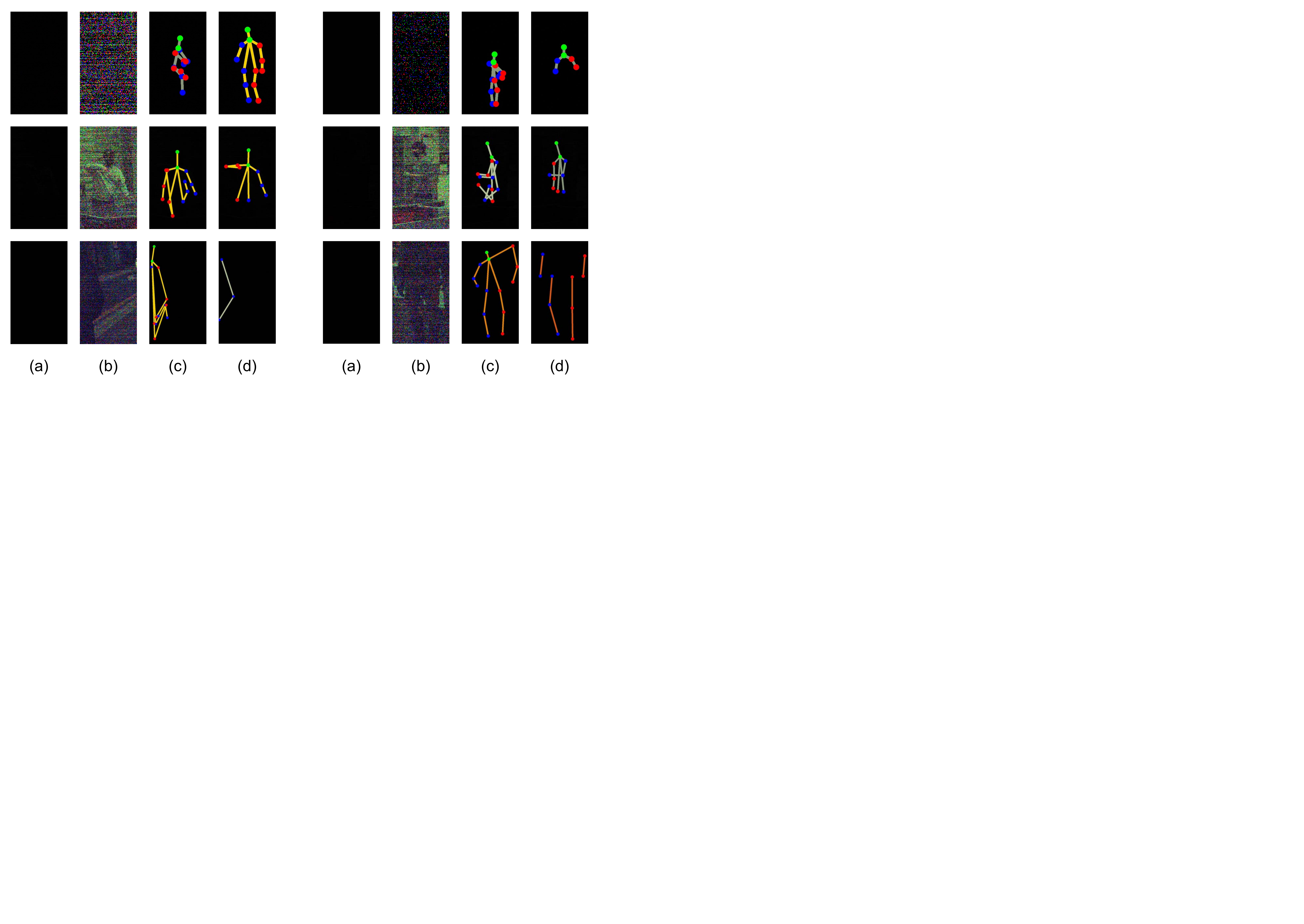} }
    \vspace{-4mm}
\caption{
Failure cases of our method. (a) Low-light images. (b) Scaled low-light images. (c) Our results. (d) Ground-truth.
} \label{fig:failure}
\vspace{-2mm}
\end{figure*}

\begin{figure*}[t]
    \centering
    \vspace{-2mm}
    \scalebox{1.0}{
    \includegraphics[width=0.85\linewidth]{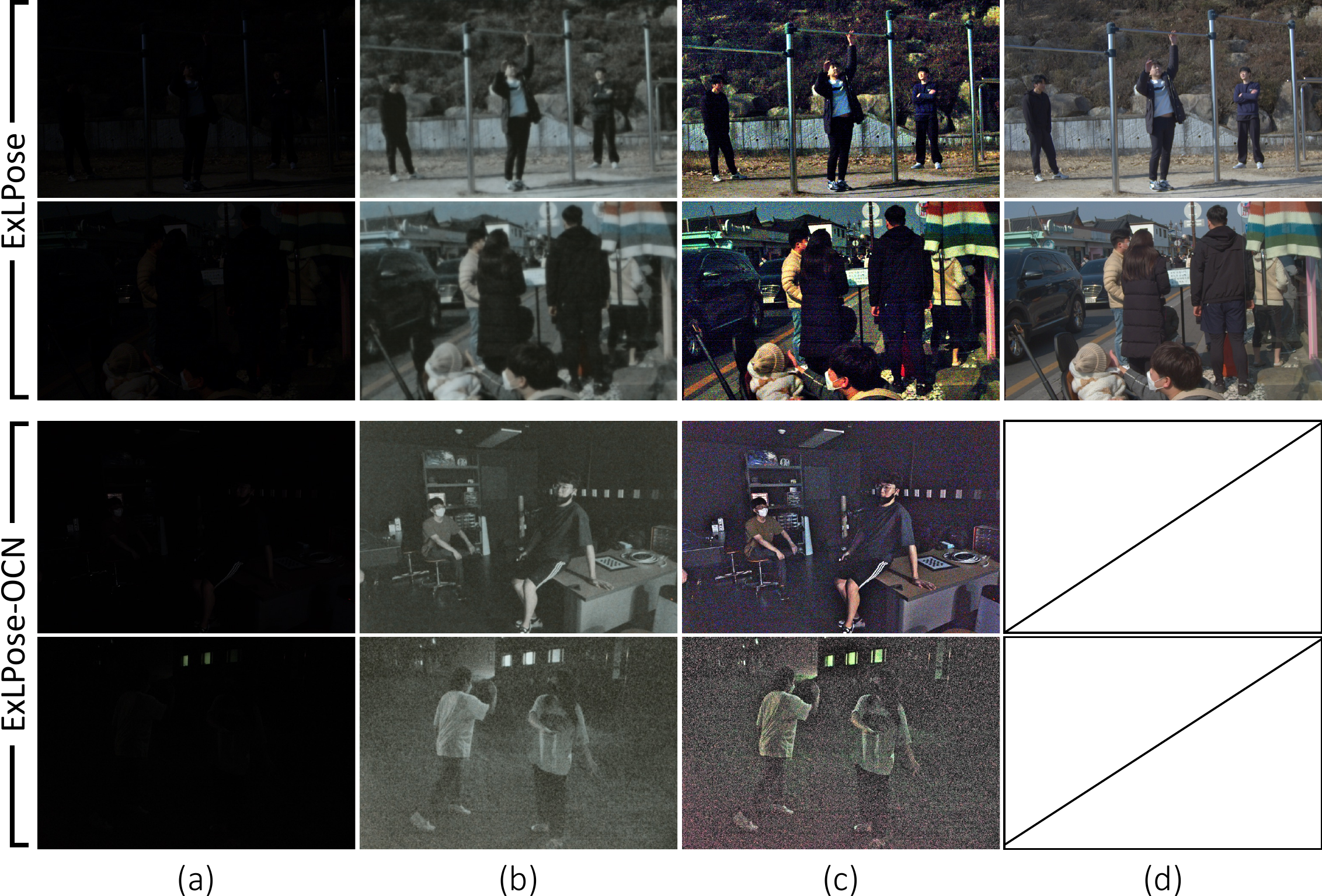} }
    \vspace{-3mm}
\caption{
Qualitative results of enhanced low-light images on the ExLPose and ExLPose-OCN datasets.
We except well-lit images of the ExLPose-OCN dataset as the dataset provides only low-light images.
(a) Low-light images. (b) LLFlow. (c) LIME. (d) Well-lit images.
} \label{fig:enhance}
\end{figure*}

\begin{figure*}[t]
    \centering
    \vspace{-2mm}
    \scalebox{0.95}{
    \includegraphics[width=\linewidth]{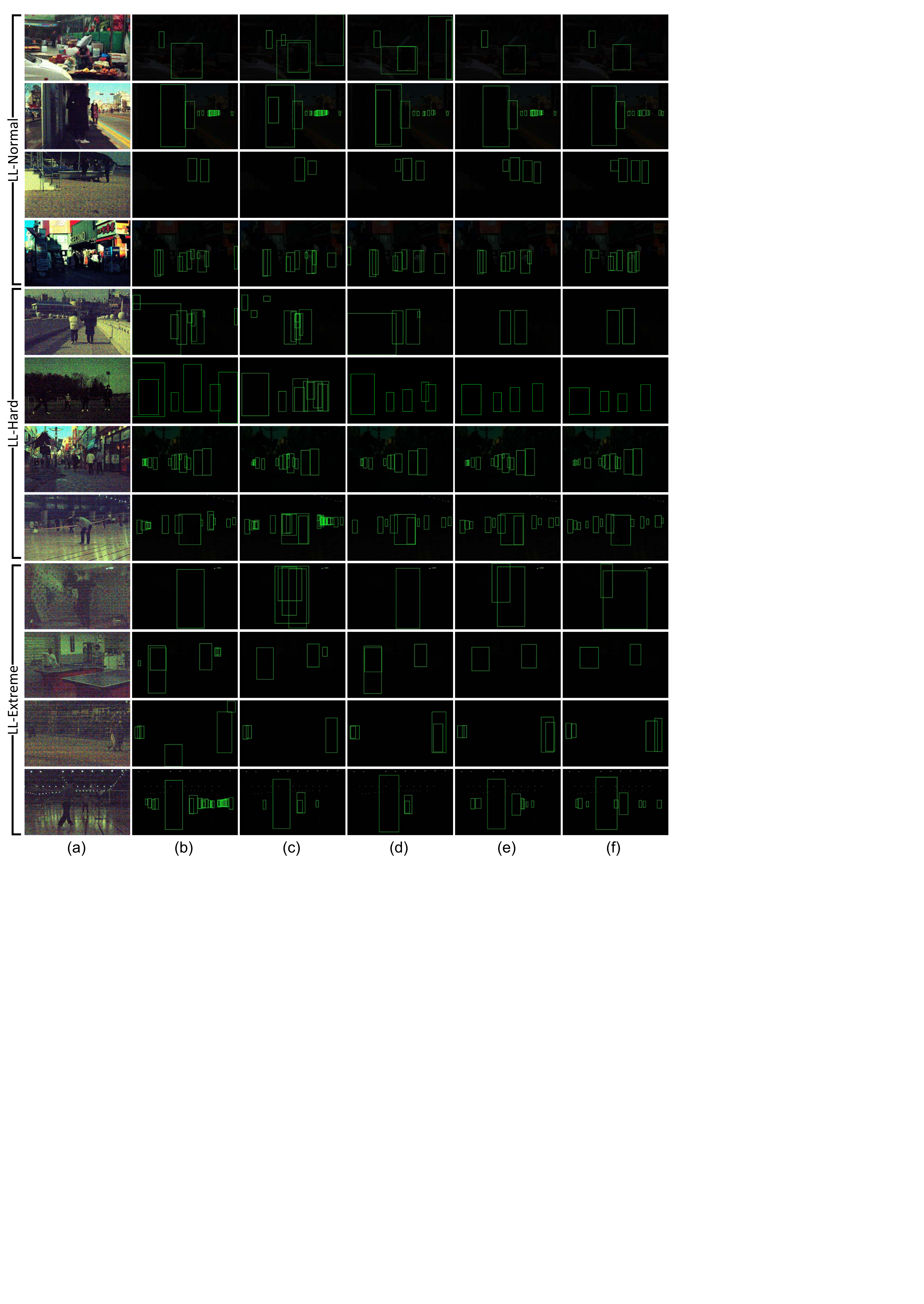} }
    \vspace{-4mm}
\caption{
Qualitative results for the person detection on the ExLPose dataset. 
Predicted boxes and labels are visualized on corresponding low-light images.
(a) Scaled low-light images. (b) Baseline-all. (c) DANN. (d) LIME + Baseline-all. (e) Ours. (f) Ground-truth.
} \label{fig:detection_qual}
\vspace{-6mm}
\end{figure*}

\begin{figure*}[t]
    \centering
    \vspace{-2mm}
    \scalebox{0.95}{
    \includegraphics[width=\linewidth]{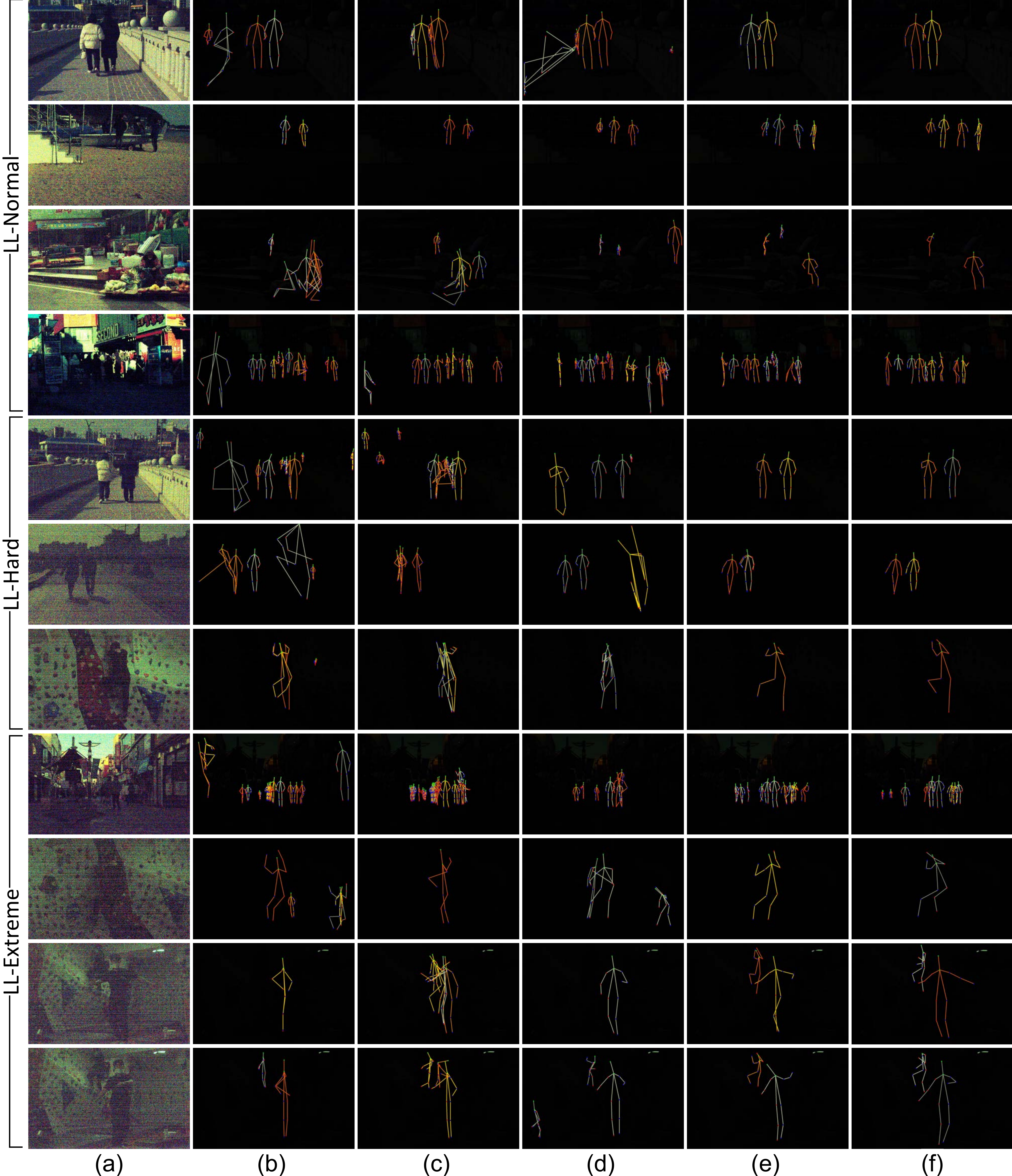} }
    \vspace{-2mm}
\caption{
Qualitative results for multi-person pose estimation on the ExLPose dataset. 
Predicted poses and labels are visualized on corresponding low-light images.
(a) Scaled low-light images. (b) Baseline-all. (c) DANN. (d) LIME + Baseline-all. (e) Ours. (f) Ground-truth.
} \label{fig:multipose_qual}
\vspace{-6mm}
\end{figure*}